%% file: main.tex
\documentclass[runningheads]{llncs}

\usepackage{eccv}

\usepackage{eccvabbrv}

\usepackage{graphicx}
\usepackage{booktabs}

\usepackage[accsupp]{axessibility}  % Improves PDF readability for those with disabilities.

\usepackage[pagebackref,breaklinks,colorlinks,citecolor=eccvblue]{hyperref}
\usepackage{orcidlink}

\def\etal{{\em et al.}}
\def\etc{{\em etc.}}

\begin{document}

\title{MirrorGaussian: Reflecting 3D Gaussians for Reconstructing Mirror Reflections} 

\titlerunning{MirrorGaussian}

% \author{Jiayue Liu\inst{1}\orcidlink{0000-1111-2222-3333} \and
% Xiao Tang\inst{2}\orcidlink{1111-2222-3333-4444} \and
% Freeman Cheng\inst{3}\orcidlink{2222--3333-4444-5555}}
\author{Jiayue Liu\inst{1}\textsuperscript{*} \and Xiao Tang\inst{2}\textsuperscript{*} \and 
Freeman Cheng\inst{3} \and Roy Yang\inst{2}
 \and Zhihao Li\inst{2}\textsuperscript{\textdagger} \and \\ Jianzhuang Liu\inst{4} \and 
Yi Huang\inst{4} \and Jiaqi Lin\inst{1} \and
Shiyong Liu\inst{2} \and Xiaofei Wu\inst{2} \\
Songcen Xu\inst{2} \and Chun Yuan\inst{1}\textsuperscript{\textdagger} }

\authorrunning{Jiayue Liu et al.}

% \institute{Tsinghua University \and
% Springer Heidelberg, Tiergartenstr.~17, 69121 Heidelberg, Germany

% \institute{ \inst{1} Tsinghua University \quad \quad \inst{2} Huawei Noah’s Ark Lab \\ 
% \inst{3} University of Toronto \quad \quad \inst{4} University of Chinese Academy of Sciences}

\institute{\quad\inst{1}Tsinghua University  \quad \quad\inst{2}Huawei Noah’s Ark Lab\\
\quad\inst{3}University of Toronto \quad \quad\inst{4}University of Chinese Academy of Sciences}

\maketitle
% \vspace{-10pt}
% \maketitle 
\begin{figure*}[htbp]
	\scriptsize
	\centering
\begin{minipage}[t]{0.2435\textwidth}	\centerline{\includegraphics[width=3.055cm]{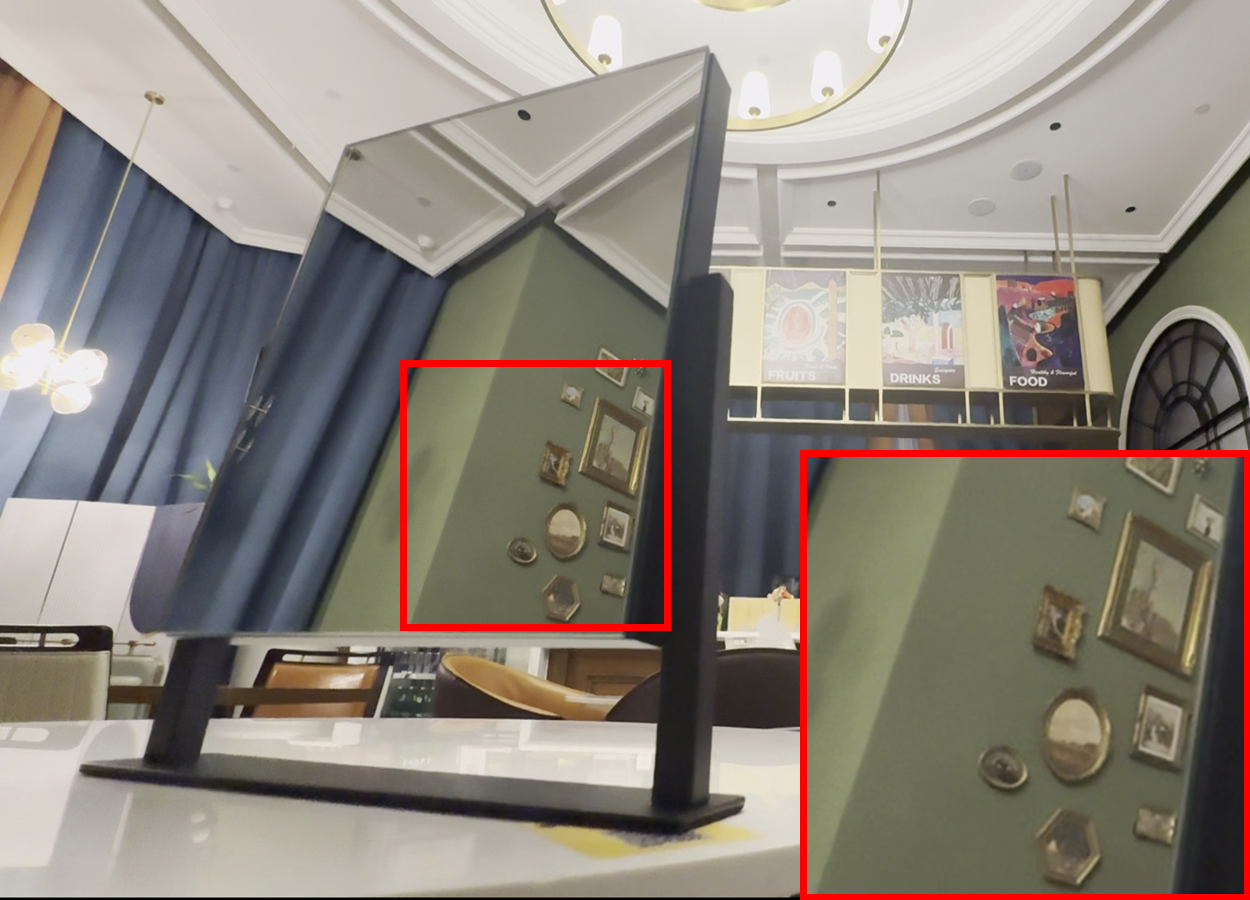}}
 \centerline{PSNR$\uparrow$/LPIPS$\downarrow$/FPS$\uparrow$}
	\centerline{Ground Truth}
 \centerline{\quad}
	\end{minipage}
\begin{minipage}[t]{0.2435\textwidth}	\centerline{\includegraphics[width=3.055cm]{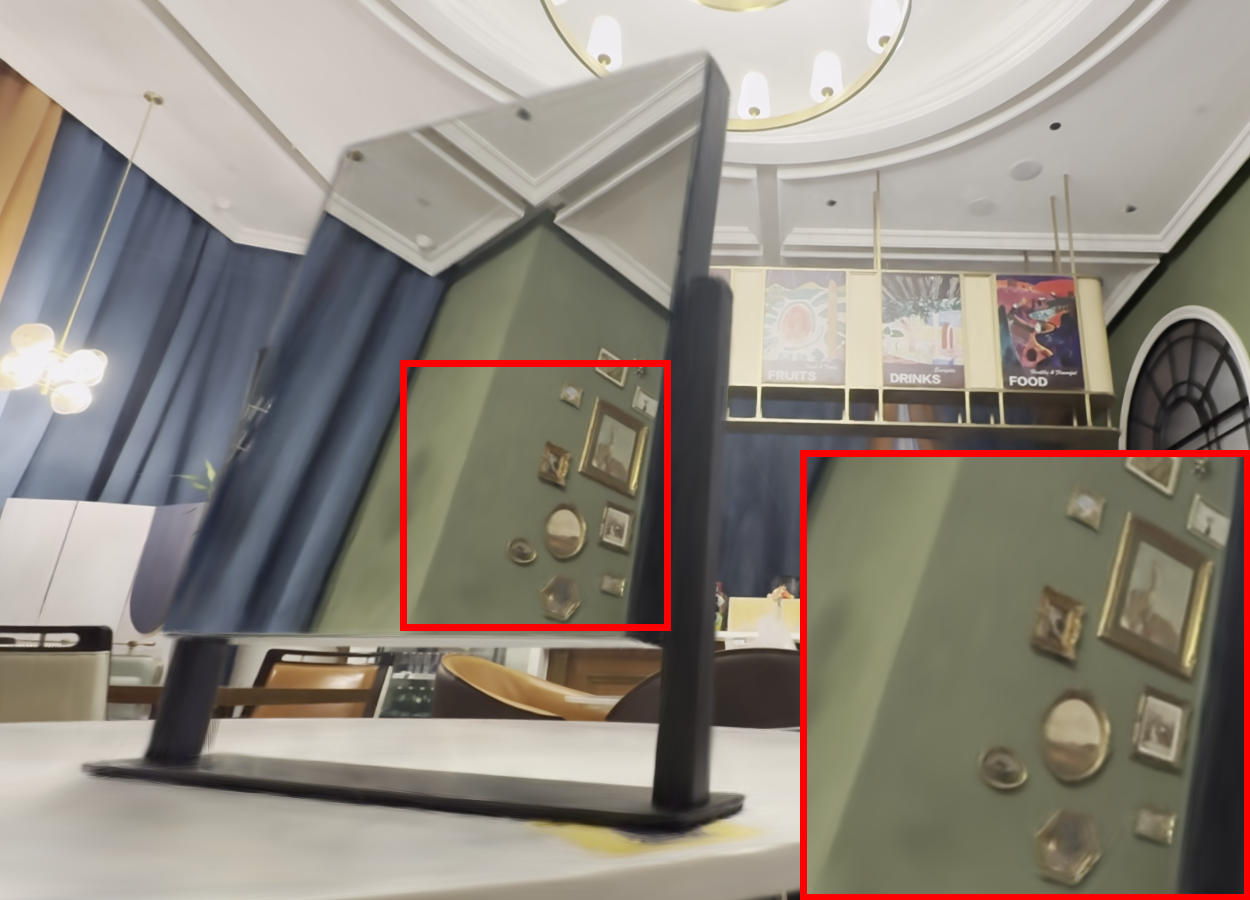}}
 \centerline{\textbf{27.78}/\textbf{0.059}/154}
 \centerline{MirrorGaussian (Ours)}
 \centerline{\quad}
	\end{minipage}
\begin{minipage}[t]{0.2435\textwidth}	\centerline{\includegraphics[width=3.055cm]{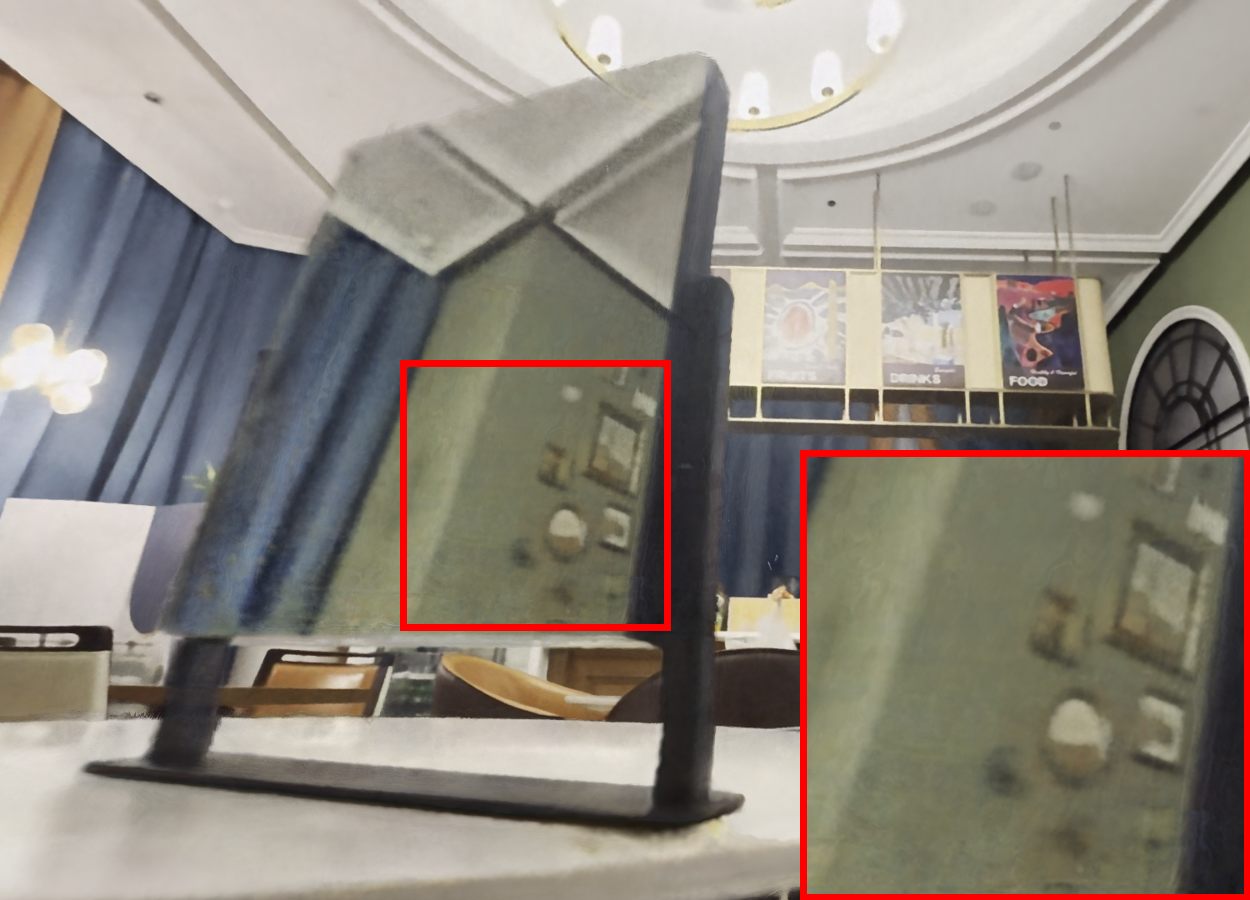}}
 \centerline{26.54/0.132/0.012}
 \centerline{MS-NeRF (CVPR'23)}
 \centerline{\quad}
	\end{minipage}
\begin{minipage}[t]{0.2435\textwidth}	\centerline{\includegraphics[width=3.055cm]{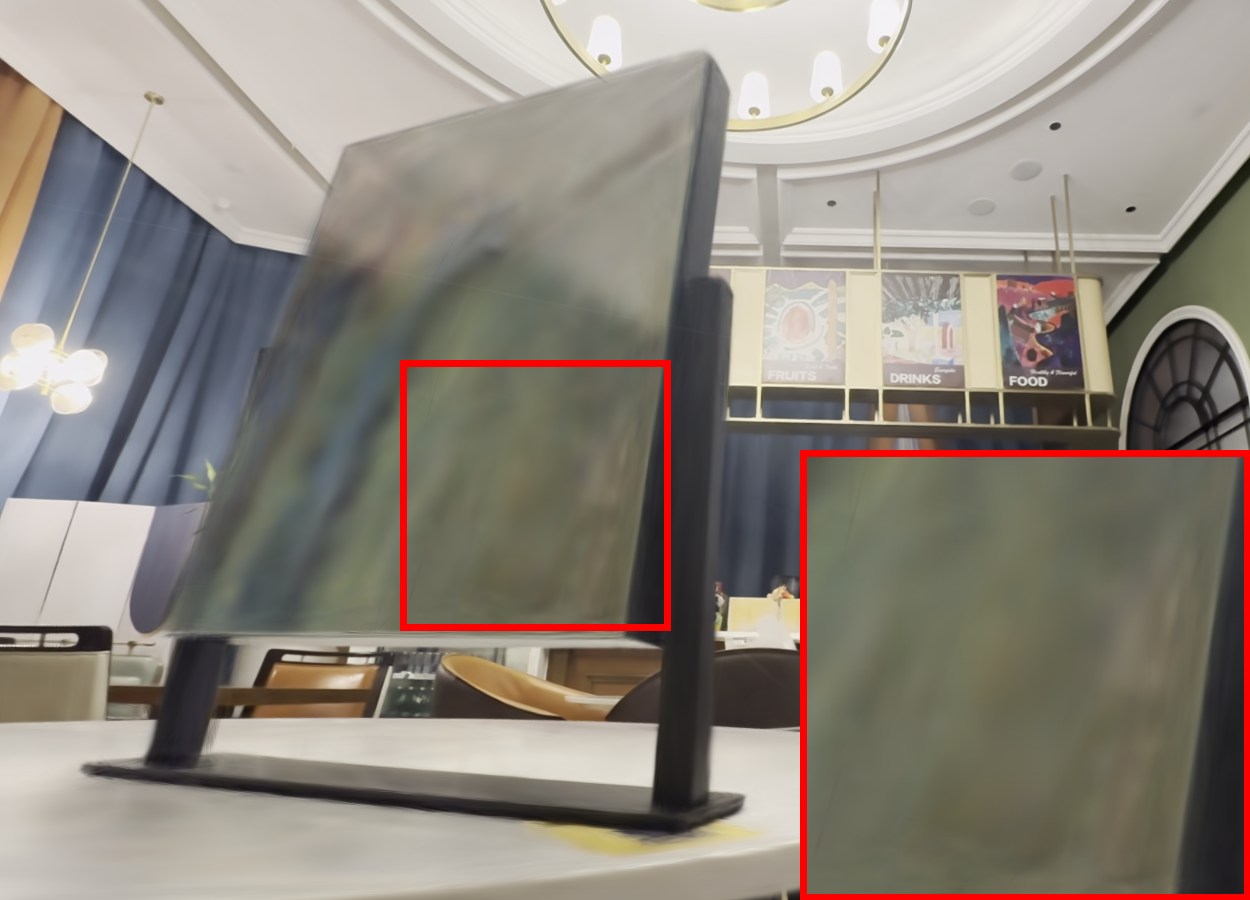}}
 \centerline{25.03/0.133/\textbf{271}}
			\centerline{3DGS (ToG'23)}
   \centerline{\quad}
	\end{minipage}
 \hfill
\begin{minipage}[t]{0.2435\textwidth}	\centerline{\includegraphics[width=3.055cm]{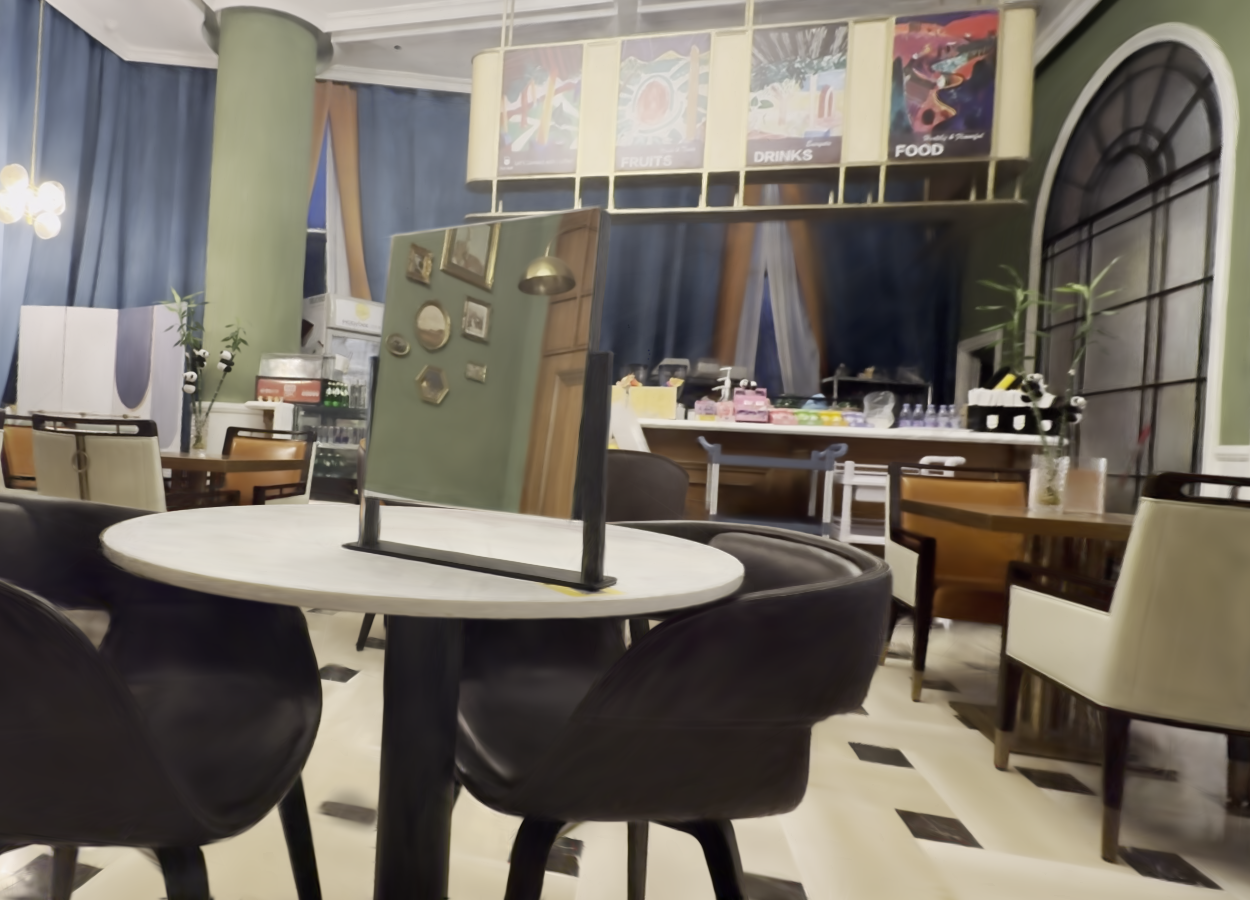}}
 		\centerline{Original}
	\end{minipage}
\begin{minipage}[t]{0.2435\textwidth}	\centerline{\includegraphics[width=3.055cm]{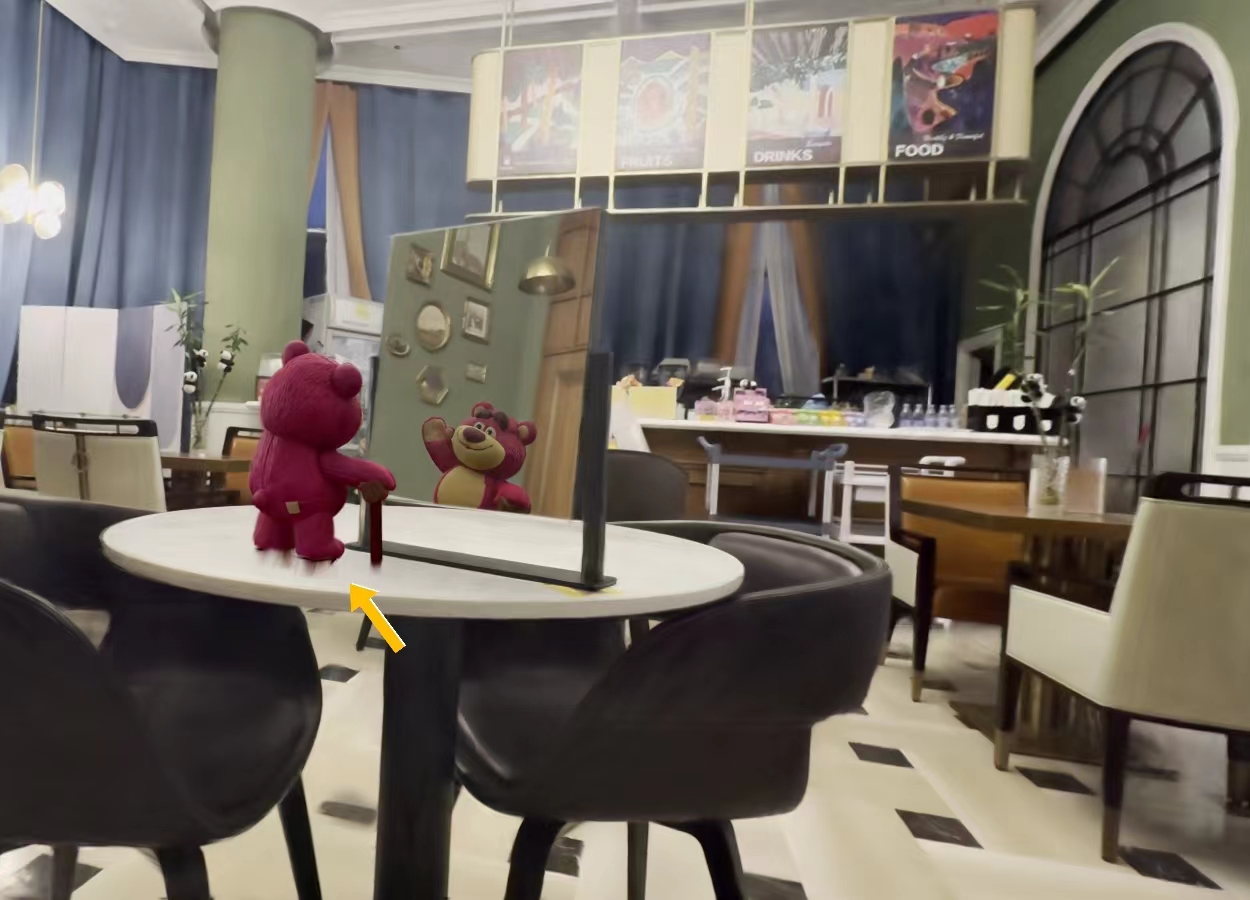}}
 \centerline{Adding an Object}
	\end{minipage}
\begin{minipage}[t]{0.2435\textwidth}	\centerline{\includegraphics[width=3.055cm]{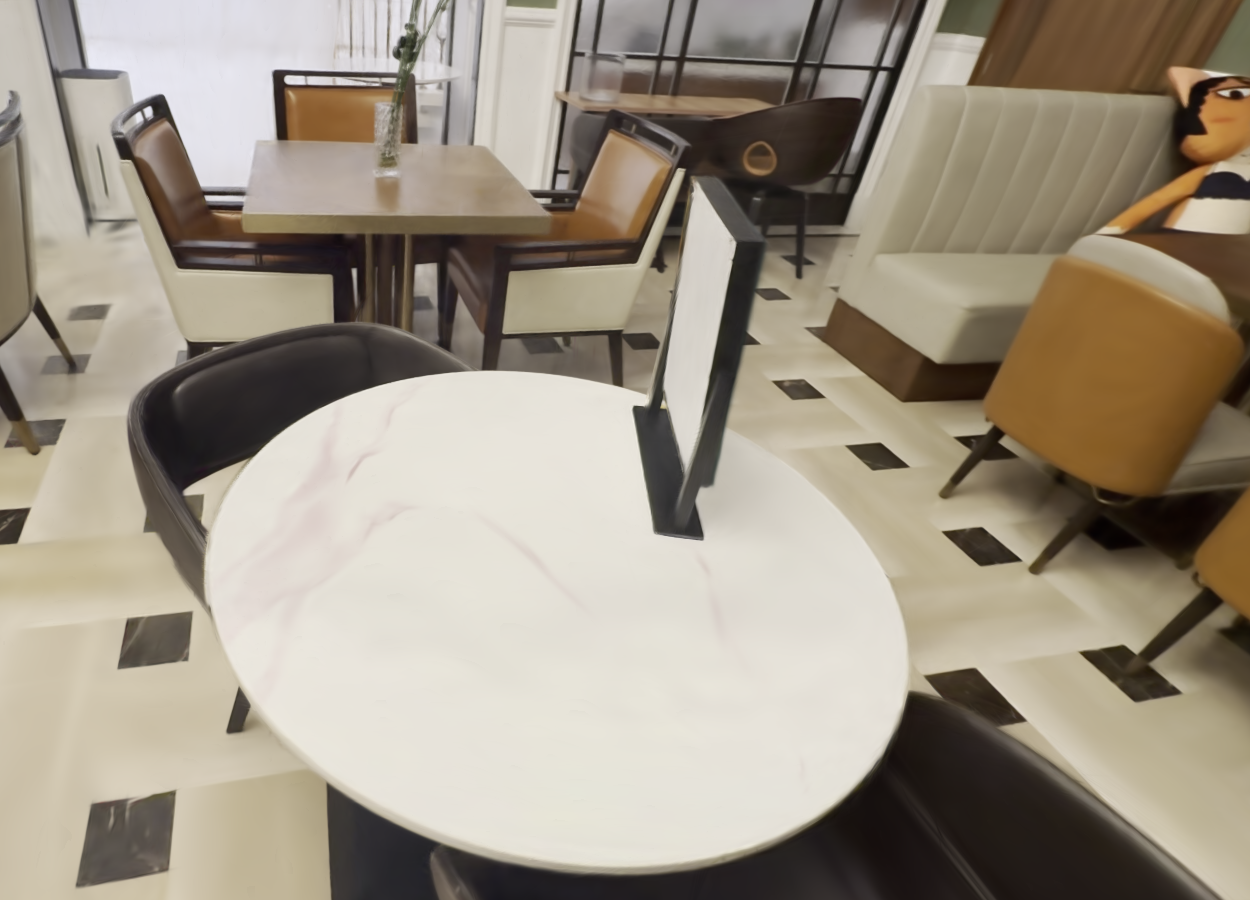}}
 \centerline{Original}
	\end{minipage}
\begin{minipage}[t]{0.2435\textwidth}	\centerline{\includegraphics[width=3.055cm]{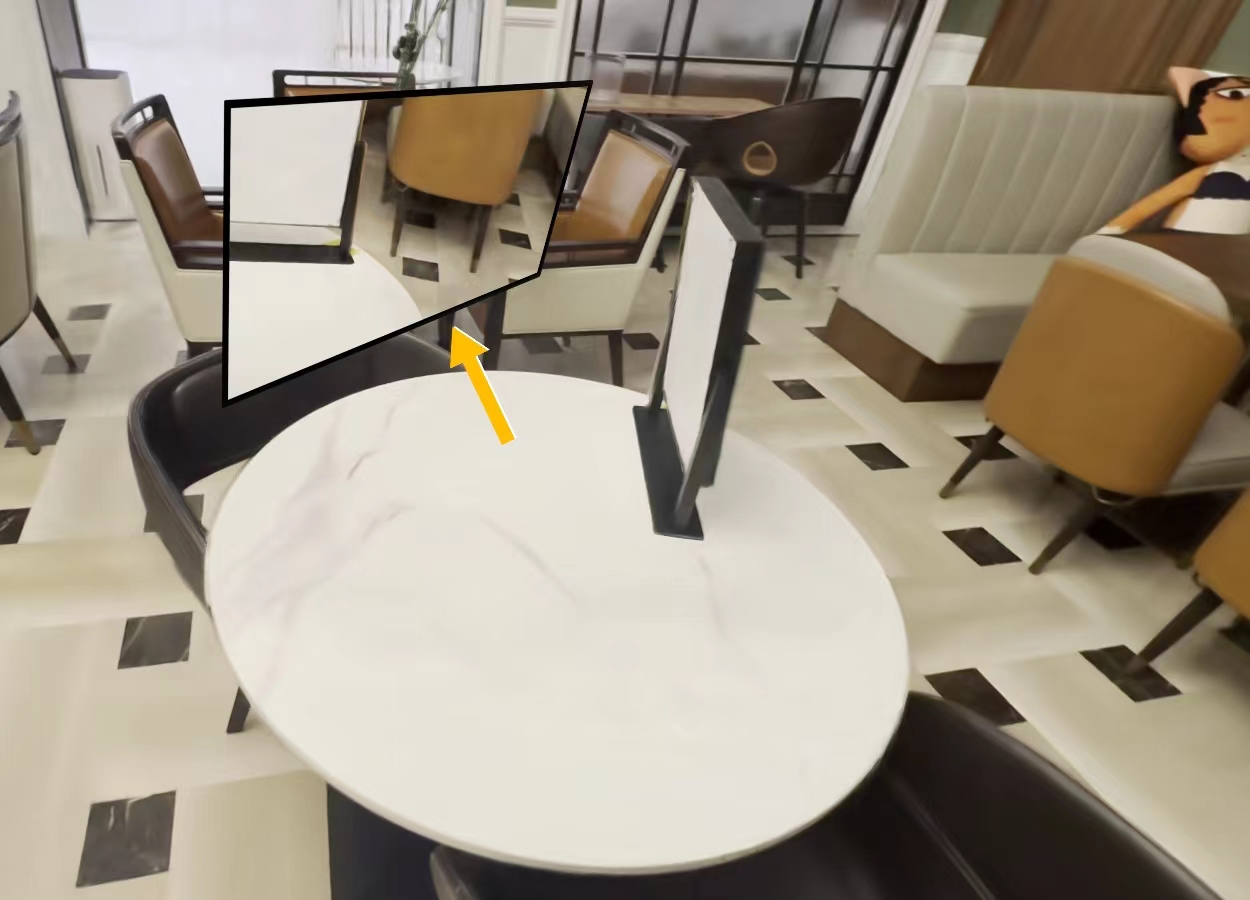}}
		\centerline{Adding a Mirror}
	\end{minipage}
\caption{The ground truth and renderings of our MirrorGaussian and state-of-the-art methods in the \textit{Coffee-House} scene from our dataset.
%MS-NeRF~\cite{yin2023multi} can reconstruct mirror reflections, while it produces blurry rendering results at low frame rates. 3D Gaussian Splatting~\cite{kerbl20233d} achieves real-time rendering, but fails to reconstruct mirror reflections.
Compared to existing methods, our MirrorGaussian achieves both high-quality and real-time rendering (top row), and empowers scene editing such as inserting new objects and mirrors (bottom row).}
	\label{fig: teaser} 
    \vspace{-1.0cm}
\end{figure*}
% \begin{center}
% \centering 
% \includegraphics[width=1\linewidth]{Figures/pipeline.pdf} 
% \captionof{figure}{ Teaser Figure: Examples of reconstruction by different inversion methods. While Null-Text Inversion requires fine-tuning, the other three methods do not. Dual-Schedule Inversion achieves excellent performance without fine-tuning. } \label{fig: teaser} 
% \end{center}
\newcommand{\tx}[1]{{\color[rgb]{0.7,0.2,0.3}\textbf{TX:}#1}}
\newcommand{\roy}[1]{{\color[rgb]{0.8,0.2,0.8}\textbf{Roy:}#1}}
\renewcommand{\thefootnote}{}
\footnotetext{\footnotesize{The work was done during an internship at Huawei.}}
\footnotetext{\footnotesize{$*$} \footnotesize{Equal contribution} \quad \footnotesize{\textdagger} \footnotesize{Corresponding authors}}

\begin{abstract}

% The recent advancements in 3D Gaussian Splatting (3DGS) have set new benchmarks in novel view synthesis, achieving state-of-the-art results and facilitating real-time rendering. However, despite its success on many datasets, 3DGS struggles with modeling mirror reflections due to multi-view inconsistency. Addressing this challenge, we present MirrorGaussian, an innovative neural rendering framework that extends 3DGS to accurately render scenes with mirrors, enabling photo-realistic novel view synthesis and real-time performance. Leveraging the principle of light reflection, our approach utilizes both the original Gaussian point cloud and its mirrored counterpart, along with a mirror mask derived from rasterization, to create authentic depictions of real and virtual images. We introduce a three-stage pipeline for precise mirror position determination and joint parameter optimization. Our experiments show that MirrorGaussian outperforms existing methods in rendering quality, training time, and rendering speed, while also supporting various scene manipulation applications.

3D Gaussian Splatting showcases notable advancements in photo-realistic and real-time novel view synthesis. However, it faces challenges in modeling mirror reflections, which exhibit substantial appearance variations from different viewpoints.
To tackle this problem, we present MirrorGaussian, the first method for mirror scene reconstruction with real-time rendering based on 3D Gaussian Splatting.
The key insight is grounded on the mirror symmetry between the real-world space and the virtual mirror space.
We introduce an intuitive dual-rendering strategy that enables differentiable rasterization of both the real-world 3D Gaussians and the mirrored counterpart obtained by reflecting the former about the mirror plane.
All 3D Gaussians are jointly optimized with the mirror plane in an end-to-end framework.
MirrorGaussian achieves high-quality and real-time rendering in scenes with mirrors, empowering scene editing like adding new mirrors and objects.
Comprehensive experiments on multiple datasets demonstrate that our approach significantly outperforms existing methods, achieving state-of-the-art results.
Project page: \href{https://mirror-gaussian.github.io/}{https://mirror-gaussian.github.io/}.

\end{abstract}

\input{1.introduction}
\input{2.related_works}

\input{3.preliminaries}
\input{4.method}
\input{5.experiments}

\input{6.conclusion}

\bibliographystyle{splncs04}
\bibliography{main}
\end{document}

%% file: 1.introduction.tex
\section{Introduction}
\label{sec:intro}
Reconstructing 3D scenes from multi-view images for photo-realistic rendering is a fundamental problem in computer vision and graphics, with wide-ranging applications in cinematography, simulation, virtual and augmented reality, \etc~
%
% Recent advancements in 3D reconstruction technologies, 
% such as Neural Radiance Fields (NeRF) \cite{mildenhall2020nerf} and 3D Gaussian Splatting (3DGS) \cite{kerbl20233d} have demonstrated remarkable outcomes. 
A notable breakthrough in this area is Neural Radiance Field (NeRF) \cite{mildenhall2020nerf} introduced in 2020. NeRF intensively samples points via ray marching, utilizes multi-layer perceptrons (MLPs) to estimate density and view-dependent colors for each point, and then adopts volume rendering to generate photo-realistic rendering results.
More recently, 3D Gaussian Splatting (3DGS) \cite{kerbl20233d} emerges and gains a lot of attention~\cite{lu2023scaffold,lin2024vastgaussian,yu2023mip,jiang2023gaussianshader,ye2023gaussian, hu2024semantic, cen2023segment}. 3DGS employs point clouds to represent scenes, where each point has opacity and anisotropic 3D Gaussian properties to model its shape, and uses a set of coefficients of spherical harmonic (SH) functions to model its view-dependent colors.
Through efficient point-based differentiable rasterization, 3DGS achieves high-quality and real-time rendering simultaneously.

However, both NeRF and 3DGS have limitations in accurately reconstructing scenes containing mirrors, due to their inherent reliance on multi-view consistency. 
Mirrors have highly specular reflections, which causes inconsistencies between the front and back views of the mirror.
As a result, the appearance of the same point can vary significantly when observed from different viewpoints.
It is challenging to model such appearance variation through MLPs or SH functions~\cite{liu2023nero, yin2023multi, jiang2023gaussianshader, yang2024spec}.
This challenge leads to a situation where mirror rendering is often blurry (see Figure~\ref{fig: teaser}) or floaters appear behind the mirror.
Some researchers propose NeRF-based approaches to reconstruct scenes with mirror reflections~\cite{yin2023multi,zeng2023mirror}. MS-NeRF~\cite{yin2023multi} introduces a multi-space scheme that constructs the scene with multiple sub-spaces, and utilizes a gate MLP that controls the visibility of a certain sub-space to obtain the final rendering results. Mirror-NeRF \cite{zeng2023mirror} and TraM-NeRF~\cite{van2023tram} trace reflected rays physically in the scene to optimize a unified NeRF model.
While these NeRF-based approaches demonstrate some success in reconstructing mirror reflections, they still face limitations. Specifically, they require expensive point sampling and MLP queries for rendering, leading to slow optimization and rendering. Since these methods are designed for ray marching, they cannot be directly adopted by rasterization to achieve fast rendering.
% intro 提纲
% novel view neural rendering 
% highly-view dependent
% 几何 慢 训练时间长 ……
% 优势：实时 质量高 训练时间短

% In this paper, we introduce MirrorGaussian, a novel point-based rendering framework extended from 3DGS, to achieve photo-realistic novel view synthesis and real-time rendering in scenes with mirrors. 
In this paper, we introduce MirrorGaussian, which integrates an explicit point-cloud-based representation extended from 3DGS for 3D reconstruction with mirror reflections.
MirrorGaussian achieves photo-realistic and real-time novel view synthesis with a rasterization-based dual-rendering strategy.
The fundamental insight is grounded in the principle of mirror reflection, which indicates the mirror symmetry between the real-world scene outside the mirror and the reflected scene inside the virtual mirror space. 
Drawing from this principle, we intuitively propose a dual-rendering strategy, which utilizes both the real-world image and its corresponding mirror image to synthesize novel views.
The real-world image is synthesized from the real-world 3DGS point cloud. The mirror image, on the other hand, is derived by reflecting the real-world 3DGS point cloud across the mirror plane.
To acquire the mirror plane equation, we first obtain a rough estimate from the sparse point cloud generated using Structure-from-Motion (SfM)~\cite{schonberger2016structure}, and then jointly optimize it with 3D Gaussians to achieve higher accuracy.
With the real-world 3D Gaussians and its mirrored counterpart, we apply the point-based rasterization~\cite{kerbl20233d} to get both the real-world and the mirror images. We then compose them using the corresponding mirror mask
%, which is also derived from the rendering process, 
to get the final rendering result.
% (Current datasets typically lack views from behind mirrors or feature sparse captures with restricted heights and perspectives. Addressing this, we have developed a dataset specifically tailored for the evaluation of 360-degree and multi-perspective rendering of scenes incorporating mirrors.) 
%
% To obtain the mirror mask from any free viewpoints, we require the dense 3D segmentation of the mirror itself.
To obtain the mirror mask from arbitrary viewpoints, we need to locate the mirror points in 3D space.
Thus, we augment 3D Gaussians with an extra mirror label, which indicates whether a point belongs to the mirror surface or not. 
%We can then adopt the point-based rasterization~\cite{kerbl20233d} for the mirror label to generate a rendered mirror mask from any view.
We can then render these 3D Gaussians to generate a mirror mask from any viewpoint.
% % and supervise it using the ground-truth mask.
% Built upon these strategies, we propose a three-stage pipeline for end-to-end optimization. We first optimize a vanilla 3DGS point cloud, and then start optimizing the mirror plane equation using the ground-truth mirror mask while fixing other parameters. Finally, we jointly optimize the 3DGS point cloud with mirror labels and the mirror plane equation to get the final results. 
% % Moreover, to avoid multi-viewpoint discrepancies, we adapt the RGB rendering formula, so that Gaussians categorized as mirror components are excluded from contributing to the color rendering process.
% %

We conduct comprehensive experiments on several datasets and demonstrate that MirrorGaussian significantly outperforms existing methods both quantitatively and qualitatively. 
We facilitate free-viewpoint navigation with real-time performance, thanks to the efficient point-based rasterization and the freedom from any neural network.
Since MirrorGaussian utilizes explicit point clouds to represent scenes, it additionally enables various applications in scene editing, such as integration of new objects into the scene, and placing a new mirror, \etc~In summary, the main contributions of this paper are threefold:

\begin{itemize}
\item We present MirrorGaussian, the first method that achieves high-fidelity reconstruction and real-time rendering of scenes containing mirrors, empowering various applications in scene editing. 
\item We propose a novel representation of scenes with mirrors for point-based rendering, which contains both the real-world 3DGS point cloud and its mirrored counterpart, obtained by reflecting the former across the mirror plane.
\item We introduce intuitive dual-rendering strategy that enables differentiable rasterization of both the real-world and the mirrored 3D Gaussians. This strategy facilitates the generation of plausible images in mirrors while maintaining efficiency of optimization and rendering.
% \item A three-stage pipeline that enables end-to-end optimization. Our pipeline maintains high-quality reconstruction of real-world scenes, while significantly improves the mirror reflection rendering.
\end{itemize}

%% file: 2.related_works.tex
\section{Related Work}

\subsubsection{Novel View Synthesis (NVS).}
% Traditional multiview reconstruction methods mainly rely on the multiview consistency of 3D points to build correspondences and estimate the depth values on different views，which is not sufficiently robust to noise and outliers. Many recent works try to introduce neural networks to estimate correspondences for the Multiview Stereo (MVS) task. The outputs of Structure-from-Motion (SfM) methods, \ie sparse point cloud and camera poses, are often used as preliminary inputs for deep-learning-based reconstruction.
% referred to NeRO
%
Given a set of calibrated images capturing a 3D scene, NVS aims to generate photo-realistic images from new viewpoints~\cite{sitzmann2019deepvoxels,shih20203d,chan2023generative,wiles2020synsin}.
A remarkable breakthrough in this field is NeRF~\cite{mildenhall2020nerf}.
% NeRF is an coordinate-based neural representation for photorealistic view synthesis that uses an MLP that maps from any continuous input 3D coordinate to the geometry and appearance of the scene at that location. 
NeRF, as an implict representation of the scene, utilizes MLPs to estimate density and view-dependent colors for intensively sampled points via ray tracing, and then leverages volume rendering to generate novel views. While NeRF can produce photo-realistic rendering results, it suffers from high computation demands on intensive point sampling and expensive MLP queries. This leads to inefficient optimization and slow rendering, making NeRF unsuitable for interactive applications.
%
% NeRF methods are constrained by the need for dense sampling of points along rays and deep network queries, resulting in slow training and inference speed.
% Despite numerous subsequent efforts employing various techniques to optimize efficiency\cite{cao2023real, chen2022tensorf, chen2023mobilenerf, fridovich2022plenoxels, hedman2021baking, muller2022instant, reiser2021kilonerf, reiser2023merf}, it remains challenging for NeRF-based methods to balance between quality and efficiency.
% As NeRF becomes a popular 3D representation for photo-realistic novel-view synthesis in recent years, many variants are proposed to improve quality, increase speed, extend to dynamic scenes, and so on.
% Various extension
Further advancements have been made to tackle this problem\cite{cao2023real, chen2022tensorf, chen2023mobilenerf, fridovich2022plenoxels, hedman2021baking, muller2022instant, reiser2021kilonerf, reiser2023merf}. However, it still remains challenging for NeRF-based methods to balance between quality and efficiency.
Concurrently, point-based rasterization methods have shown impressive results, providing an appealing mix of rendering quality and computational efficiency~\cite{ruckert2022adop, franke2023vet, franke2024trips, kerbl20233d, kopanas2021point}.
A representative work is 3DGS~\cite{kerbl20233d}, which starts gaining much attention. 3DGS explicitly represents a 3D scene with a set of anisotropic 3D Gaussians points, and each point contains opacity and a set of SH coefficients for modeling view-dependent colors. By applying a hardware-accelerated point-based rasterizer instead of computationally intensive ray tracing, 3DGS achieves both high-quality rendering and real-time rendering at the same time. 
% The recent expressive paradigm of 3D Gaussian Splatting \cite{kerbl20233d} represents a scene with a set of anisotropic 3D Gaussians that inherit the differential properties of volumetric representation. \roy{expanded 3DGS introduction here} It differs from the traditional point-based rendering methods where the 3D points are fixed in size. 3D Gaussian points vary in size, orientation, opacity and color represented by Spherical Harmonics. By splatting each 3D point onto the 2D image space, it takes advantage of a hardware-accelerated tile-based rasterizer \cite{zwicker2002ewa} instead of resource-intensive ray-marching, thereby achieving real-time rendering while retaining high quality. 
% As 3DGS becomes a popular 3D representation for photorealistic novel-view synthesis, it is adopted in many applications such as avatar creation \cite{Zielonka2023Drivable3D, jena2023splatarmor}, autonomous driving \cite{zhou2023drivinggaussian, yan2023streetgaussians}, 3D content generation \cite{tang2023dreamgaussian, yi2023gaussiandreamer}, and so on. 
Lately, there have been a lot of work on further enhancing the quality of 3DGS~\cite{lu2023scaffold,lin2024vastgaussian,lee2024deblurring,yu2023mip,yan2023multi}. Scaffold-GS~\cite{lu2023scaffold} proposes an anchor-based approach to distribute 3D Gaussians, which delivers higher rendering quality. Mip-Splatting~\cite{yu2023mip} identifies the aliasing problem in 3DGS and introduces a 3D smoothing filter and a 2D Mip filter to produce alias-free results.
However, both NeRF-based and 3DGS-based methods still struggle with reconstructing scenes with mirrors. Mirrors have highly specular reflections, which can vary significantly from different viewpoints. It remains challenging for existing methods to model such variation using MLPs or SH functions. 

\subsubsection{Reflection Reconstruction.} % \roy{rewriting this section}}%Referred to PPT
Some studies have started to tackle the reflection reconstruction via NeRF~\cite{srinivasan2021nerv,yao2022neilf,liu2023nero,guo2022nerfren} or point-based rendering~\cite{kopanas2022neural,jiang2023gaussianshader,liang2023gs,shi2023gir}.
One common approach is decomposing diffuse colors and specular colors from the objects by physically-based rendering (PBR). 
% These methods typically require to accurately reconstruct the geometry of the object and calculate the specular colors from a differentiable environment lighting map. 
NeRO~\cite{liu2023nero} first reconstructs the geometry of the object using NeuS~\cite{wang2021neus}, and then recovers the environment lights and the materials of the object via PBR to calculate specular colors. GaussianShader~\cite{jiang2023gaussianshader} extends 3DGS to extract the normal of the reflective object from the shortest axis directions of 3D Gaussians, and similarly optimizes an environment lighting map for rendering reflective appearances.
These methods mostly rely on optimizing the environment lighting map; while it might be sufficient for object-level reconstruction, it is often too rough to model reflections in the mirror.
%
% Despite the significant advancement in point-based rendering, similar to NERF, vanilla 3DGS also struggles with handling strong specular reflections due to the violation of multi-view consistency. This is because the scenes observed on different sides of the reflective surfaces are different and under such circumstances, a foggy virtual world containing lots of artifacts tends to be created through training, which fails to accurately represent the reflected content.
% Our MirrorGaussian is the first approach to deal with such a intractable situation in 3DGS. \roy{should delete the above sentence.}
% Although there has not been sufficient research on improving scenes with reflective surfaces in terms of 3DGS, numerous NeRF based methods have been proposed. 
%NeRF 3DGS 没法处理反射 
% One common approach is to respect the multi-view consistency assumption. To this extent, studies have proposed different ways to properly model the real and the virtual world separately and render the final image as a weighted combination of diffuse and reflective elements, which our work draws inspiration on.
Another direction is to model the real-world and the reflected elements separately, and then combine them with appropriate weights to render the final image.
%
% Ref-NeRF \cite{verbin2022ref} structures view-dependent outgoing radiance with reflected view direction and a collection of spatially-varying scene properties. 
Ref-NeRF \cite{verbin2022ref} reparameterizes NeRF's color MLP to predict reflection colors from reflected view directions about estimated normal vectors.
% BakedSDF \cite{yariv2023bakedsdf} 
% merges reflected view radiance fields with camera view-based radiance fields, leveraging a hash grid backbone to speed up training. 
%
NeRF-ReN \cite{guo2022nerfren} models both a real-world radiance field and a reflected radiance field. The images rendered from these two fields are further blended using an optimizable weight.
%
% Similarly, UniSDF \cite{wang2023unisdf} optimizes a real-world radiance field additionally with a the reflected radiance field, and seamlessly combine them using a weight field to compute the final color.
UniSDF \cite{wang2023unisdf} estimates the SDF of the scene to predict surface normals, and similarly optimizes two radiance fields and blending weights. The reflective radiance field takes the reflected view direction derived from the surface normal as input to better capture reflective colors.
As for point-based rendering, Kopanas~\etal~\cite{kopanas2022neural} reconstruct both real-world and reflected point clouds, and use an MLP to model curved reflectors' trace of reflection inside the object to reconstruct reflections.
While these methods achieve separate optimization for real-world and reflective elements, they fail to consider the correspondence between the two.
This oversight may limit the ability to reconstruct high-fidelity details from mirror reflections, thus compromising overall rendering quality.
% Neural Point Catacaustics \cite{kopanas2022neural} utilize dual point cloud representation designed for curved reflectors, but it relies on the MVS reconstruction for initialization and couldn't enable real time rendering.

% No need to cite
% Besides these, Spec-Gaussian \cite{yang2024spec} utilizes an anisotropic spherical Gaussian (ASG) appearance field instead of SH for modeling the view-dependent appearance of each 3D Gaussian.

% Other than above scene-level reconstruction methods, several works \cite{liu2023nero, jin2023tensoir, gao2023relightable} attempt to decompose a scene's view-dependent appearance into material and lighting properties. They leverage the Bidirectional Reflectance Distribution Function (BRDF) to deduce geometric shapes, ambient lighting conditions, and surface materials. Although considering both direct and indirect lighting, they fall short in scenarios characterized by highly specular mirror reflections because the environment map is not suitable for modeling such reflections.

% Several inverse rendering methods leverage the Bidirectional Reflectance Distribution Function (BRDF) to deduce geometric shapes, ambient lighting conditions, and surface materials. 

%object level: inverse rendering (envrionment map)：慢 几何 
%参考文献：gs 5篇 

\subsubsection{Mirror Reflection Reconstruction.}
There are some NeRF-based methods that specifically address the challenge of reconstructing mirror reflections. MS-NeRF \cite{yin2023multi} innovates by decomposing the scene into multiple subspaces, each represented by a neural feature field. To render the scene, they sample points in these subspaces to get multiple feature maps. These feature maps are passed through a decoder MLP to get multiple rendered images. A gate MLP is then applied to compose these images into the final rendering result.
% and introduces a lightweight module attached to NeRF's output layer. These subspaces are then coherently merged through a learnable gating layer as an enhancement to NeRF-backbone networks. 
% Mirror-NeRF \cite{zeng2023mirror} takes a more physically principled approach by tracing the paths of incoming and outgoing rays, thereby increasing the number of sampling points. The normal directions and reflection probabilities are additionally learned to determine the weights for color composition.
% \tx{TODO} Mirror-NeRF \cite{zeng2023mirror} takes a more physically principled approach by tracing the paths of incoming and outgoing rays. They additionally estimation the normal and reflection probabilities are additionally learned to determine the weights for color composition.
Mirror-NeRF \cite{zeng2023mirror} optimizes a unified NeRF by tracing rays on reflective surfaces using the estimated normals. Additionally, it estimates the reflection probabilities of surface to blend the colors of camera rays and reflected rays, then synthesizes the final image.
% thereby increasing the number of sampling points. The normal directions and reflection probabilities are additionally learned to determine the weights for color composition.
%
% Similarly, TraM-NeRF~\cite{van2023tram} also learns a unified radiance fields through ray tracing. However, it aims to reduce the number of samplings needed through more efficient strategies for importance sampling and transmittance computation, which effectively lowers the computational cost of the evaluation of the neural networks.
TraM-NeRF~\cite{van2023tram}, a concurrent work, employs a similar ray-tracing strategy and introduces a radiance estimator that combines volume and reflected radiance integration to reduce the number of sampled points along rays.

% However, it aims to reduce the number of samplings needed through more efficient strategies for importance sampling and transmittance computation, which effectively lowers the computational cost of the evaluation of the neural networks.

% These methodologies are impeded by slow training and rendering speeds due to inefficient ray traing.% any others?

However, these methods still rely on inefficient ray marching and expensive MLP queries. This results in heavy computation loads, making them impractical for interactive usage.
In contrast, our MirrorGaussian excels in superior reconstruction quality and fast rendering, positioning it as a promising solution for various applications in scene editing.

% Compared to existing methods, MirrorGaussian excels in two key aspects. First, it achieves superior reconstruction quality for scenes with mirrors. Second, MirrorGaussian facilitates efficient optimization and real-time rendering, making it suitable for various interactive applications.

%% file: 3.preliminaries.tex
\section{Preliminaries}
\label{sec:preliminaries}
% Starting from a set of SfM points
3DGS represents the scene with $Q$ anisotropic 3D Gaussian primitives $\{\mathcal{G}_i | i=1,...,Q \}$:
\begin{equation}
\mathcal{G}_i(x) = e^{-\frac{1}{2}(x-\mu_i)^T \Sigma_i^{-1}(x-\mu_i)},
\end{equation}
where $\mu_i \in \mathbb{R}^{3}$ denotes $\mathcal{G}_i$'s center (mean) and $\Sigma_i \in \mathbb{R}^{3 \times 3}$ denotes its 3D covariance matrix defined in the world space. 
To maintain its positive semi-definiteness, $\Sigma_i$ is decomposed as an ellipsoid using a scaling matrix $S_i \in \mathbb{R}^{3 \times 3}$ and a rotation matrix $R_i \in \mathbb{R}^{3 \times 3}$:
\begin{equation}
\Sigma_i = R_i S_i S^T_i R^T_i.
\label{eq:sigma_decomposition}
\end{equation}
Each 3D Gaussian $\mathcal{G}_i$ is also characterized with opacity $\alpha_i \in \left[0, 1\right]$ and spherical harmonic coefficients $SHs$ for view-dependent colors $c_i(d)$, where $d$ denotes the view direction from the camera. 

During the rendering process, 3D Gaussians $\mathcal{G}_i(x)$ are projected to 2D Gaussians $\mathcal{G'}_i(x)$ with $\mu'_i \in \mathbb{R}^{2}$ and $\Sigma'_i \in \mathbb{R}^{2 \times 2}$ via EWA Splatting~\cite{zwicker2002ewa, kerbl20233d}, where $\mu'_i$ and $\Sigma'_i$ denotes the center (mean) and the 2D covariance matrix of $\mathcal{G}'_i$ in the image space, respectively. $\{\mathcal{G}'_i\}$ are then assigned to different tiles, sorted and alpha-blended into a rendered image in a point-based volume rendering manner. The color $C$ of a pixel $p$ is computed via alpha blending:
\begin{gather}
\label{eq:accumulation}
    C = \sum_{i \in N} c_i(d) \alpha_i \mathcal{G}'_i(p) T_i, \quad T_i = \prod_{j=1}^{i-1}(1-\alpha_j \mathcal{G}'_j(p)), \\
    \mathcal{G}'_i(p) = e^{-\frac{1}{2}(p-\mu'_i)^T \Sigma'^{-1}_i(p-\mu'_i)}, \notag
\end{gather}
where $N$ denotes the set of ordered 2D Gaussians overlapping the pixel. Leveraging the differentiable tile-based rasterizer, all attributes of 3D Gaussians are optimized end-to-end efficiently via the combination of the $\mathcal{L}_1$ and D-SSIM losses:
\begin{equation}
    \mathcal{L}_c = (1-\lambda)\mathcal{L}_1 + \lambda \mathcal{L_{D-SSIM}},
\end{equation}
where $\lambda$ is a balance parameter. This process is interleaved with adaptive point densification and pruning to better represent the scene\cite{kerbl20233d}.

%refer to vastgaussian mipsplatting scaffoldGS
%Each of them is parameterized by its center(mean) $\mu \in \mathbb{R}^{3 \times 1}$, 3D covariance matrix defined in world space $\Sigma \in \mathbb{R}^{3 \times 3}$, opacity $\alpha \in [0,1]$, and spherical harmonic coefficients for view-dependent colors. 

%% file: 4.method.tex
\section{Method}

\begin{figure}[tb]
    \centering
    \includegraphics[width=\textwidth]{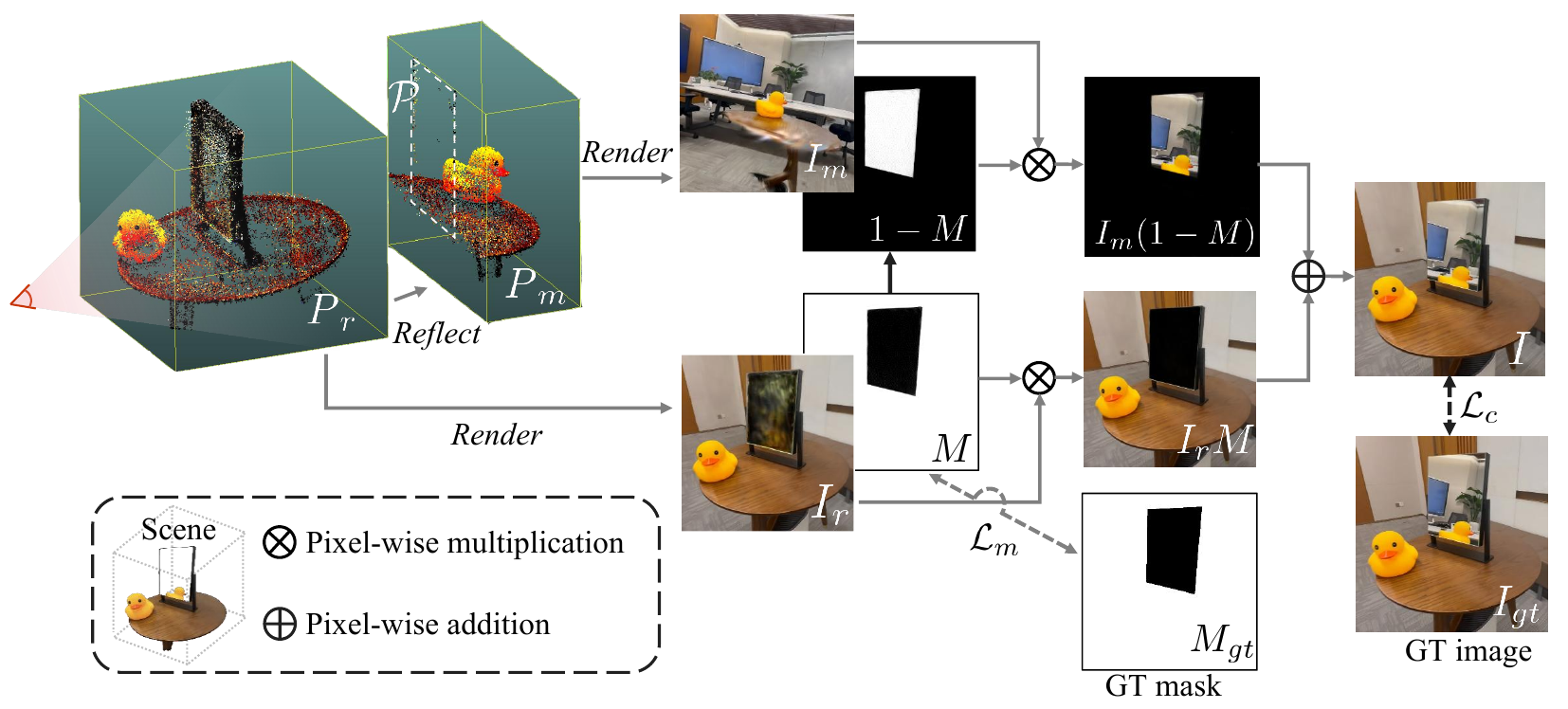}
    \caption{\textbf{Overview of MirrorGaussian.} MirrorGaussian is grounded on the mirror symmetry between the real-world scene and its counterpart in the mirror. We first reflect the 3D Gaussians $P_r$ about the mirror plane $\mathcal{P}$ to obtain its mirrored counterpart $P_m$. Then, we rasterize $P_r$ to get the real-world image $I_r$ and the mirror mask $M$, and rasterize $P_m$ to get the mirror image $I_m$. The final image $I$ is composited by $I_r$ and $I_m$ using $M$. $I$ and $M$ are supervised by the captured image $I_{gt}$ and its annotated mirror mask $M_{gt}$, respectively. Note that for the sake of visual simplicity, $P_r$ and $P_m$ have been cropped.
    %
    % . To enable precise flipping, we devise a training method to obtain an accurate plane position. Then, we rasterize $P_r$ and $P_m$ to get the rendered image $I_r$ and mask $M$ for the real-world objects, and the rendered image $I_m$ for the virtual reflection by introducinga category attribute to indicate whether a point is on the surface of the mirror. The final image $I$ is composited by $I_r$ and $I_m$ using $M$.
    }
    \label{fig:pipeline}
\end{figure}

% We utilize both the original Gaussian point cloud and its flipped counterpart, positioned across the mirror plane, for rendering authentic real-world image outside the mirror and the corresponding virtual image inside it. Then these two renderings are fused using a mirror mask derived from the rendering process.

% In this work, we develop a pipeline MirrorGaussian utilizing 3DGS for the high-fidelity reconstruction and real-time rendering of scenes with mirrors. Our approach takes as input multi-view RGB images, the corresponding camera poses, and mirror masks. 

% The overview of MirrorGaussian is illustrated in Fig. \ref{fig:pipeline}. MirrorGaussian takes as input a set of calibrated images capturing a static scene containing a mirror from various viewpoint, along with the corresponding mirror masks and a sparse point cloud obtained from Structure from Motion (SfM) \cite{schonberger2016structure}.
The overview of MirrorGaussian is illustrated in Fig. \ref{fig:pipeline}. MirrorGaussian reconstructs a static scene containing a mirror from multiple-view images, along with their corresponding camera poses and mirror masks, and a sparse point cloud obtained from Structure from Motion (SfM) \cite{schonberger2016structure}.
% As illustrated in \cref{fig:pipeline}, given a novel view, MirrorGaussian performs the following steps: 
%
% (1) \textbf{Dual-Rendering}: Initially, we rasterize the original Gaussians for capturing the scene external to the mirror, and subsequently, rasterize the flipped Gaussians in front of the mirror to simulate the virtual reflections internal to the mirror. 
%
% (2) \textbf{Scene Composition}: We generate the mirror's mask during rasterization by category attribute, facilitating an integrated representation of both the real and reflected worlds.
%
% Next we describe each component in detail.
%
% The core idea of MirrorGaussian is derived from the principle of light reflection, where real-world objects and their mirrored counterparts exhibit symmetrical positioning about the mirror plane. 
% %
% We render images of both the scene outside the mirror and the scene reflected within the mirror, and subsequently stitch these two images together. Given that objects inside the mirror maintain a symmetrical relationship with real-world objects, we simulate the virtual reflections inside the mirror by rasterizing the flipped Gaussians in front of the mirror (\cref{sec: flipping}). To facilitate the flipping process, we designed a two-stage training method to obtain the mirror's precise location (\cref{sec: mirrPos}). To integrate the rendered images of the real world and the mirror reflection, we determine the mirror's mask during the rasterization process by introducing a categorical attribute and modifying the RGB rendering formula (\cref{sec: composition}).
%
The core idea of MirrorGaussian is derived from the principle of mirror reflection: a scene in the real world and its reflected counterpart in the virtual mirror space are mirror-symmetrical about the mirror plane. 
Leveraging this principle, we propose a dual-rendering strategy, which renders both the real-world image $I_r$ outside the mirror and the corresponding mirror image $I_m$, and fuses them to get the final rendering result.
Based on our dual-rendering strategy, we further propose a three-stage pipeline for end-to-end optimization of reconstructing scenes containing mirrors.
First, we start by optimizing 3DGS in an unmodified manner to get the real-world 3D Gaussians $P_r$. Next, we reflect $P_r$ to get the mirrored 3D Gaussians $P_m$ in the mirror space, and optimize the mirror plane equation via our dual-rendering strategy while fixing other parameters. Finally, we augment $P_r$ with a mirror label for generating the mirror mask in any viewpoints. We jointly optimize 3D Gaussians with the mirror label and the mirror plane equation to achieve high-fidelity reconstruction with mirror reflection.

% First, we flip the real-world 3DGS point cloud $P_r$ in front of the mirror about the mirror plane to get the mirrored 3DGS point cloud $P_m$ in the virtual mirror space (\cref{sec: flipping}). 
%
% To facilitate the flipping process, we designed a two-stage training method to obtain the mirror plane (\cref{sec: mirrPos}). Next, we follow~\cite{kerbl20233d} to render both  $P_r$ and $P_m$ to get the real-world image $I_r$ and the mirror image $I_m$.
% %
% Finally, we fuse $I_r$ and $I_m$ using the rendered mirror's mask $M$ during the optimization process, to obtain our final rendering result $I$ (\cref{sec: composition}).

In the following subsections, we first explain how to reflect $P_r$ to get mirrored 3D Gaussians $P_m$ given the mirror plane equation (\cref{sec: flipping}). 
Next, we present a method that obtains a rough mirror plane equation from the sparse SfM point cloud, and jointly optimize it with 3D Gaussians via our dual-rendering strategy (\cref{sec: mirrPos}). 
Finally, we describe how to optimize the 3D mirror mask, enabling high-quality rendering of mirror reflections from arbitrary viewpoints (\cref{sec: composition}).

\subsection{3D Gaussians Reflection about the Mirror Plane}
\label{sec: flipping}

%We parameterize the plane using a quaternion $p$ to express the plane equation. 
% Assume we have determined the plane equation, 
% In order to render reflections within a mirror, 
This subsection explains how to reflect the 3D Gaussians $P_r$ across the mirror plane $\mathcal{P}$ (the acquisition of $\mathcal{P}$ is detailed in~\cref{sec: mirrPos}) to get the mirrored 3D Gaussians $P_m$ for modeling the virtual mirror space.
%
% This mirroring operation requires adjusting Gaussians in three key aspects: center $\mu$, covariance $\Sigma$ and color $c$.  
We parameterize the mirror plane as $\mathcal{P}(x) = \langle n, x \rangle + b = 0, x \in \mathbb{R}^3$, where $n \in \mathbb{R}^3$ denotes the normal vector of $\mathcal{P}$ and $b \in \mathbb{R}^1$ denotes the negative of the plane's distance from the origin along $n$. A mirrored 3DGS point $\mathcal{\hat{G}}$ is derived from $\mathcal{G}$ by keeping its opacity $\alpha$ and scale $S$, but modifying its other three attributes: mean $\mu$, rotation $R$ and view-dependent color $c(d)$. 
%
% First, the position of the mirrored point $\mu'$ is determined by preserving the same distance from the plane as $\mu$, but situated on the opposite side of the plane. The center of the mirrored Gaussian $\mu'$ is computed as:
% \begin{equation}
%     \mu' = \mu - 2 \left(n \cdot \left( \mu - q_{plane}  \right) \right) n,
% \end{equation}
% where $q_{plane}$ denotes a point on the plane.
% First, the mean of the mirrored Gaussian point $\mu'$ is determined by preserving the same distance from the plane as $\mu$, but situated on the opposite side of the plane. The center of the mirrored Gaussian $\mu'$ is computed as:
% \begin{equation}
%     \mu' = \mu - 2 \left(n \cdot \left( \mu - q_{plane}  \right) \right) n,
% \end{equation}
% where $q_{plane}$ denotes a point on the plane.

First, the mean $\hat{\mu}$ of $\hat{\mathcal{G}}$ can be determined using the reflection function $\mathcal{F}$:
\begin{equation}
    \hat{\mu} = \mathcal{F}(\mu) =  \mu - 2 \frac{\langle n, \mu \rangle + b}{\lVert n \rVert^2} n.
\end{equation}

\begin{figure}[tb]
    \centering
    \includegraphics[width=0.9\textwidth]{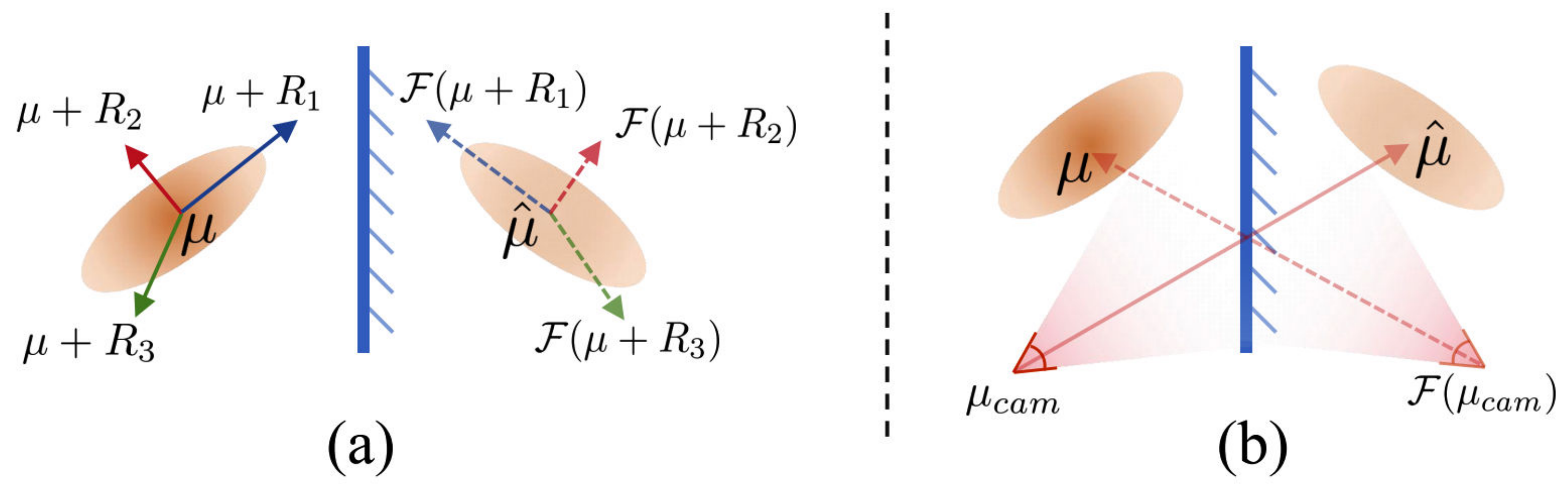}
     \caption{Reflecting a real-world 3D Gaussian $\mathcal{G}$ with mean $\mu$ and rotation $R$ about the mirror plane (blue line) to get the mirrored 3D Gaussian $\hat{\mathcal{G}}$. (a) illustrates how the mean $\hat{\mu}$ and the rotation $\hat{R}$ of the mirrored 3D Gaussian $\hat{\mathcal{G}}$ are derived using the reflection function $\mathcal{F}$.
     % Flipping Gaussians involves adjusting the covariance matrix $\Sigma$ to modify the orientation of the distribution. The covariance matrix $\Sigma$ can be decomposed into a rotation matrix $R$, where the columns represent the axes in space. 
     %
     % (b) Maintaining the spherical harmonics coefficients, we compute the reflected color by inputting the flipped viewing directions into the spherical harmonics function.
     (b) illustrates how we derive the view-dependent color of $\hat{\mathcal{G}}$. Instead of modifying the SH coefficients according to the reflection, we directly reflect the view direction for $\hat{\mathcal{G}}$ to get the view-dependent color.
     }
    \label{fig:flipping}
\end{figure}

\noindent Next, we flip $R$ across $\mathcal{P}$ to get the rotation $\hat{R}$ of $\hat{\mathcal{G}}$ (See~\cref{fig:flipping}(a)). By revisiting Eq.~\ref{eq:sigma_decomposition}, we find that $S$ is a diagonal matrix and $R$ is an orthogonal matrix that satisfies $R^T = R^{-1}$, and thus Eq.~\ref{eq:sigma_decomposition} can also be interpreted as the eigendecomposition of the matrix $\Sigma$:
\begin{equation}
\Sigma = R S S^T R^T = R S^2 R^{-1},~R^{-1} \Sigma R = S^2.
\end{equation}
Therefore, the columns of $R$, denoted as $R_1$, $R_2$ and $R_3$, are the normalized eigenvectors of $\Sigma$. 
Given that $\Sigma$ also describes the configuration of a Gaussian ellipsoid~\cite{kerbl20233d}, $R_1$, $R_2$ and $R_3$ represent the directions of the three principal axes of the Gaussian ellipsoid in the 3D space.
Based on it, $\hat{R} = (\hat{R_1}, \hat{R_2}, \hat{R_3})$ is obtained by reflecting $R_1$, $R_2$, $R_3$ over $\mathcal{P}$:
\begin{equation}
\hat{R_i} = \mathcal{F}(\mu + R_i) - \hat{\mu}, ~i = 1,2,3 .
\end{equation}   
% Flipping the Gaussian entails manipulating the directions of these axes, resulting in a transformation of the rotation matrix $R$, which in turn changes in the covariance matrix $\Sigma$.
% This transformation effectively flips the Gaussian's orientation while preserving its shape by the original scaling matrix $S$.

% \tx{TODO}Last, Since 3DGS uses spherical harmonics coefficents and the viewing direction to model view-dependent color $c$, we can compute the color $c'$ of a point reflected in the mirror by flipping the viewing direction as illustrated in \cref{fig:flipping}(b) while keeping the position of the point $\mu$ and the spherical harmonic coefficients unchanged.

Last but not least, the view-dependent colors $c(d)$ should also be flipped to get $\hat{c}(d)$ in the mirror space. 3DGS uses a set of SH coefficients to model view-dependent colors $c(d)$, but it is complex to directly compute these coefficients after reflecting. Thus, instead of computing them, we mirror $d$ when rendering $\hat{\mathcal{G}}$, as illustrated in \cref{fig:flipping}(b). Then $\hat{c}(d)$ can be derived by:
\begin{equation}
\hat{c}(d) = c(\hat{d}),~\hat{d} = \mu - \mathcal{F}(\mu_{cam}) ,
\end{equation} 
where $\mu_{cam}$ is the coordinate of the camera.

% we can compute the color $c'$ of a point reflected in the mirror by flipping the viewing direction as illustrated in \cref{fig:flipping}(b) while keeping the position of the point $\mu$ and the spherical harmonic coefficients unchanged.

% Thus, adjusting $\Sigma$ requires simply adjusting  the three columns of $R$, which in turn effectively reverses the orientation of the 3D Gaussians as shown in \cref{fig:flipping}(a).
% Flipping the Gaussian entails manipulating these axes, resulting in a transformation of the rotation matrix and, consequently, changes in the covariance matrix

% 0418_flippling_4.png

% After completing the standard rasterization process to obtain real-world rendering, to render reflections involves flipping Gaussians in front of the mirror surface while excluding those located behind it. This section explains how to flip Gaussians and determine accuarate mirror positions to facilitate this flipping.

\subsection{Mirror Plane Equation Estimation}
\label{sec: mirrPos}
\begin{figure}[tb]
    \centering
    \includegraphics[width=\textwidth]{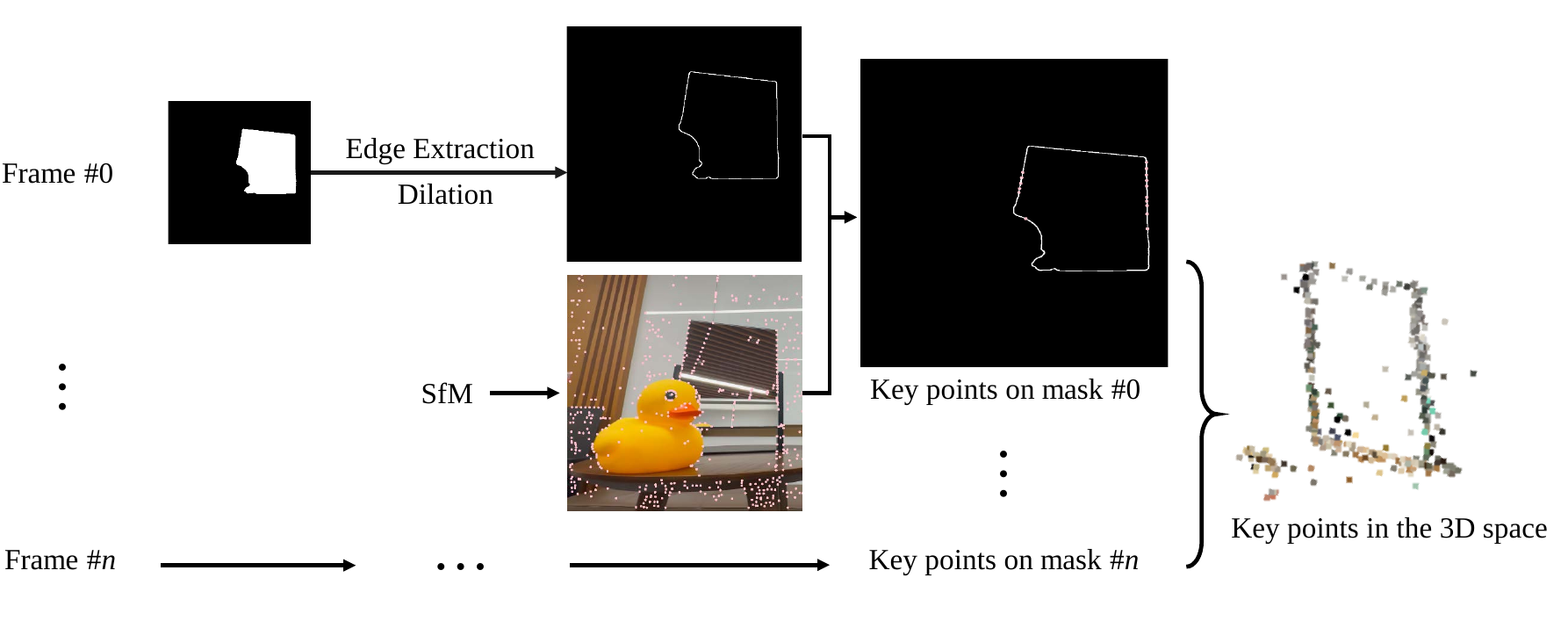}
    \caption{\textbf{Strategy for yielding the mirror's initial position.} The mirror's edges from each frame's mask are extracted using dilation and erosion. Employing SfM, corresponding 2D-3D point pairs are established, allowing reconstruction of 3D points along the mirror edges. Finally, a plane is fitted to the aggregated 3D points from all frames.}
    \label{fig:mirr pos init}
\end{figure}

% To facilitate the flipping process, we design a two-stage training method to obtain the mirror's position in space. We first get a initial position with rough accuracy by running SfM. The strategy involves the following steps as illustrated in \cref{fig:mirr pos init}. 
We need to acquire the accurate mirror plane equation via optimization to enable the mirroring process.
To reduce the difficulty of optimization, we start from a rough estimation of $\mathcal{P}$, which is derived from the input sparse point cloud generated from SfM. 
\cref{fig:mirr pos init} illustrates the procedure of obtaining this rough estimation.
Given the mirror mask of an image (derived from SAM~\cite{kirillov2023segment}), we extract the mask edges and dilate it to get the 2D mirror border. 
Next, we identify 3D points from the sparse SfM point cloud that fall on the extracted 2D border. This is achieved by leveraging the correspondences between 2D points in the images and 3D points in the point cloud, which is established by SfM.
We apply this operation to all input images to aggregate 3D points that locate on the mirror border from the SfM point cloud, then remove flying points, and finally use RANSAC~\cite{fischler1981random} to fit a plane to the 3D points as the rough estimation of the mirror plane. 

Since this mirror plane estimation may not be accurate enough, we further jointly optimize it with 3D Gaussians, which is incorporated into our three-stage optimization pipeline. In detail, in the first optimization stage, we train vanilla 3DGS for $s_1$ steps to get the real-world 3D Gaussians $P_r$. Next, we mirror $P_r$ about the roughly estimated $\mathcal{P}$ to get $P_m$. Note that we only mirror 3D Gaussian points that are in front of the mirror plane. Points behind the mirror plane are not reflected and thus have no contribution to the mirror image.
For the next $s_2$ steps in our second optimization stage, we render both $P_r$ and $P_m$ to get the real-world image $I_r$ and the mirror image $I_m$, and then fuse them together using the ground-truth mask $M_{gt}$ obtained by SAM~\cite{kirillov2023segment} to get the final rendering image $I$:
\begin{equation}
    I = I_r \odot M_{gt} + I_m \odot (1-M_{gt}).
\end{equation}
We supervise $I$ using the captured image $I_{gt}$ with the $\mathcal{L}_1$ loss and D-SSIM loss. To enhance optimization stability and ensure accurate estimation of the mirror plane equation, we only backpropagate the gradients to the parameters of the mirror plane $\mathcal{P}$. Optimizable parameters of $P_r$ are fixed during the second optimization stage. After $s_2$ steps, the mirror plane equation is further optimized in our final optimization stage, which will be detailed in~\cref{sec: composition}.

% Using SfM, we identify correspondences between image keypoints and the 3D point cloud. By dilating and eroding the mirror mask, we generate a mask around the mirror's edges. This process allows us to link the mask around the mirror borders to the 3D point cloud, from which we fit a plane using RANSAC, illustrated in \cref{fig:mirr pos init}.

\subsection{3D Mirror Mask}
\label{sec: composition}
% The main idea of MirrorGaussian revolves around a dual-rendering approach: initially, to rasterize the original Gaussians for capturing the scene external to the mirror, and subsequently, to rasterize the inverted Gaussians in front of the mirror to simulate the internal virtual reflections. To seamlessly merge these two renderings, we generate the mirror's mask during rasterization by category attribute (see below), facilitating an integrated representation of both the real and reflected worlds.

% In this section, we explain how to obtain the mirror mask in any novel views to integrate the rendered images of the real world and the mirror reflection by introducing a categorical attribute and modifying the RGB rendering formula.

% Given the real-world 3DGS point cloud and its mirrored counterpart over the mirror plane, we can render the real-world image and the mirror image
%
With our dual-rendering strategy based on the accurate mirror plane equation, we can already generate plausible renderings if the mirror masks are provided. However, to synthesize novel views, we need to acquire the mirror mask $M$ from arbitrary viewpoints. 
Thus, in the third optimization stage with $s_3$ steps, our main goal is to identify 3DGS points $\mathcal{G}$ that are located on the mirror surface. By rendering these mirror points, we can generate mirror masks from novel views.

\subsubsection{Mirror Label.}
% In order to generate mirror masks in novel views, we add an attribute $k \in [0,1]$ to each Gaussian indicating its category, with value close to zero when it is typed as mirror surface and close to one when it's not. We get the category $K$ of a pixel by differentiable rasterizing:
% \begin{equation}
%     K  = \sum_{i \in \mathcal{N}} k_i \alpha_i T_i, \quad T_i = \prod_{j=1}^{i-1}(1-\alpha_j),
% \end{equation}
%
%Inspired by~\cite{ye2023gaussian, hu2024semantic, cen2023segment}, 
We assign a label $k \in [0,1]$ for each $\mathcal{G}$ with value zero when it belongs to the mirror surface and one when it does not. We can then render the mirror mask $M$ by applying the same alpha blending from 3DGS as described in Eq.~\ref{eq:accumulation}: 
% \begin{equation}
%     K  = \sum_{i \in \mathcal{N}} k_i \alpha_i T_i, \quad T_i = \prod_{j=1}^{i-1}(1-\alpha_j),
% \end{equation}

\begin{gather}
    K = \sum_{i \in N} k_i \alpha_i {\mathcal{G}'_i}(p) T_i, \quad T_i = \prod_{j=1}^{i-1}(1-\alpha_j {\mathcal{G}'_j}(p)),
    % \mathcal{G'}(p) = e^{-\frac{1}{2}(p-\mu'_i)^T \Sigma'_i^{-1}(p-\mu'_i)} \notag
\end{gather}
where $K \in [0, 1] $ is the value of the pixel $p$ in $M$, denoting how likely $p$ belongs to the mirror. 

In the third optimization stage, we add mirror labels $k$ to all 3D Gaussians and initialize them based on the distance of 3D Gaussians' means $\mu$ to the mirror plane. We set $k=1$ when the distance exceeds a threshold $\tau = 0.1$, indicating that the point does not belong to the mirror. For the remaining 3D Gaussians, $k$ is randomly assigned between $0$ and $1$.

\subsubsection{RGB Rendering Formula with the Mirror Label.} 
% Gaussians categorized as mirror surface should not contribute to the alpha-blending process of color. This is because, as the observation viewpoint changes with varying angles and height, the same point on a mirror surface can display vastly different reflected colors. The purpose of this layer of Gaussians growing on the mirror surface is to accurately form the geometric shape of the mirror, ensuring the precision of the mirror mask rendered from novel viewpoints. 
%
%
% To ensure that Gaussians have minimal impact on pixel colors when their k values are close to zero, we modify the color rendering formula by multiplying $\alpha$ with $k$. The color $C$ of a pixel is computed as:
%
We require points with $k \approx 0$ to be distributed over the mirror surface to enable accurate mirror mask rendering. Meanwhile we do not want these points to contribute to the rendering of $I$, as they may hinder the rendering of $P_m$.
To achieve so in a differentiable way, we modify the vanilla color rendering formula in Eq.~\ref{eq:accumulation} by multiplying $\alpha$ with $k$. The color $C$ of a pixel $p$ is then computed as:
% During the course of training, a layer of gaussians with very small $k$ values gradually grow on the mirror surface. Their function is to form the geometric shape of the mirror, ensuring the accuracy of the mirror mask rendered from novel viewpoints, and guiding the flipping of point cloud. They should not contribute to the alpha-blending of color, otherwise multi-view consisitency will be violated. Therefore we multiply $\sigma$ with $k$ to ensure that Gaussian hardly affects pixel color when its $k$ is close to zero. The color $C$ of a pixel is modified as:
% \begin{equation}
%     C  = \sum_{i \in \mathcal{N}} c_i {\alpha_i}' {T_i}', \quad {T_i}' = \prod_{j=1}^{i-1}(1-{\alpha_j}'), \quad {\alpha_j}' = \alpha_j k_i.
% \end{equation}
\begin{equation}
C = \sum_{i \in N} c_i(d) {\alpha}'_i {\mathcal{G}'_i(p) {T}'_i}, \quad {T}'_i = \prod_{j=1}^{i-1}(1- {\alpha}'_j {\mathcal{G}}'_j(p)), \quad {\alpha}'_j = \alpha_j k_j.
\end{equation}

%with ${\sigma_j}' = \sigma_j k_i$.

% Then we obtain the rendering of the mirror mask. 
% With $\hat{I_r}$ and $\hat{M_r}$ representing the rendered image and mask for the real-world scene external to the mirror, and $\hat{I_m}$ and $\hat{M_r}$ indicating the rendered image and mask for the virtual scene obtained from mirrored Gaussians, the final rendered image $\hat{I}$ is combined as follows:
% To achieve high accuracy of estimating $M$, i
In the third optimization stage, we fuse the real-world image $I_r$ and the mirror image $I_m$ using the estimated mirror mask $M$ instead of $M_{gt}$ to get the final image $I$:
\begin{equation}
    I = I_r \odot M + I_m \odot (1-M).
\end{equation}
We also supervise $I$ by computing the $\mathcal{L}_1$ loss and D-SSIM loss between $I$ and the captured image $I_{gt}$. To supervise $M$, we add the mask loss $\mathcal{L}_m$, which is the $\mathcal{L}_1$ loss between $M$ and $M_{gt}$. We further add a regularization term $\mathcal{L}_d$ that encourages the 3D Gaussians belonging to the mirror surface to be close to the mirror plane:
\begin{equation}
    \mathcal{L}_d = \frac{1}{Q} \sum_{i=1}^{Q}\left[\frac{\left| \langle n, \mu \rangle + b \right|}{\lVert n \rVert}  (1-k_i)\right]^2,
\end{equation}
where $Q$ is the total number of the 3D Gaussians. This loss effectively helps the mirror points to fast converge towards the mirror plane. 

In the third optimization stage, we optimize all 3D Gaussian attributes, including the newly introduced mirror label $k$ and the mirror plane $\mathcal{P}$, but tune down the learning rate of $\mathcal{P}$ for stability. We incorporate all the loss functions mentioned above to obtain the training loss for our final optimization stage:
\begin{equation}
    \mathcal{L} = \mathcal{L}_c + \lambda_m \mathcal{L}_m + \lambda_d \mathcal{L}_d, %+ \lambda_e \mathcal{L}_e.
\end{equation}
where $\lambda_m=0.2$ and $\lambda_d=0.1$ are two balance hyperparameters. Note that the optimization losses for the first and second stages are both $\mathcal{L}_c$ only.
% During the first stage, we employ loss function akin to 3DGS\cite{kerbl20233d}, \ie $\mathcal{L}_c$. 

\subsubsection{Implementation Details.}
The training steps of the three stages $s_1$, $s_2$, and $s_3$ are 20,000, 1,000 and 9,000, respectively. We use the Adam optimizer and exponentially decay the learning rate of $\mathcal{P}$ from 1.6e-4 to 1.6e-6 in the second stage. In the third stage, the learning rate of $\mathcal{P}$ is decayed from 1.6e-6 to 1.6e-8 exponentially, and the learning rate of $k$ is set to 0.005.

% \subsection{Progressive Training}

% In the initial epochs of second-stage training, we employ the mirror mask to derive the final image \ie $\hat{I} = \hat{I}_r M + \hat{I}_v(1-M)$. After some epochs, we revert to using a rendering mask for computing the final image, \ie $\hat{I} = \hat{I}_r \hat{M}_r + \hat{I}_v(1-\hat{M}_r)$. We observe that the quality of rendered masks significantly affects the RGB learning process. This is implemented to hasten the mask's convergence during the preliminary phases of training which are unstable, thereby preventing the formation of Gaussians behind the mirror. 

% Furthermore, as training progresses into later epochs, we decay $\lambda_m$ and learn the attribute $k$ under the mere guidance of $\mathcal{L}_c$ to minimize the impact of potential mask inaccuracies.

%% file: 5.experiments.tex
\section{Experiments}

\subsection{Experiment Setup}

\begin{figure}[tb]
    \centering
    \includegraphics[width=\textwidth]{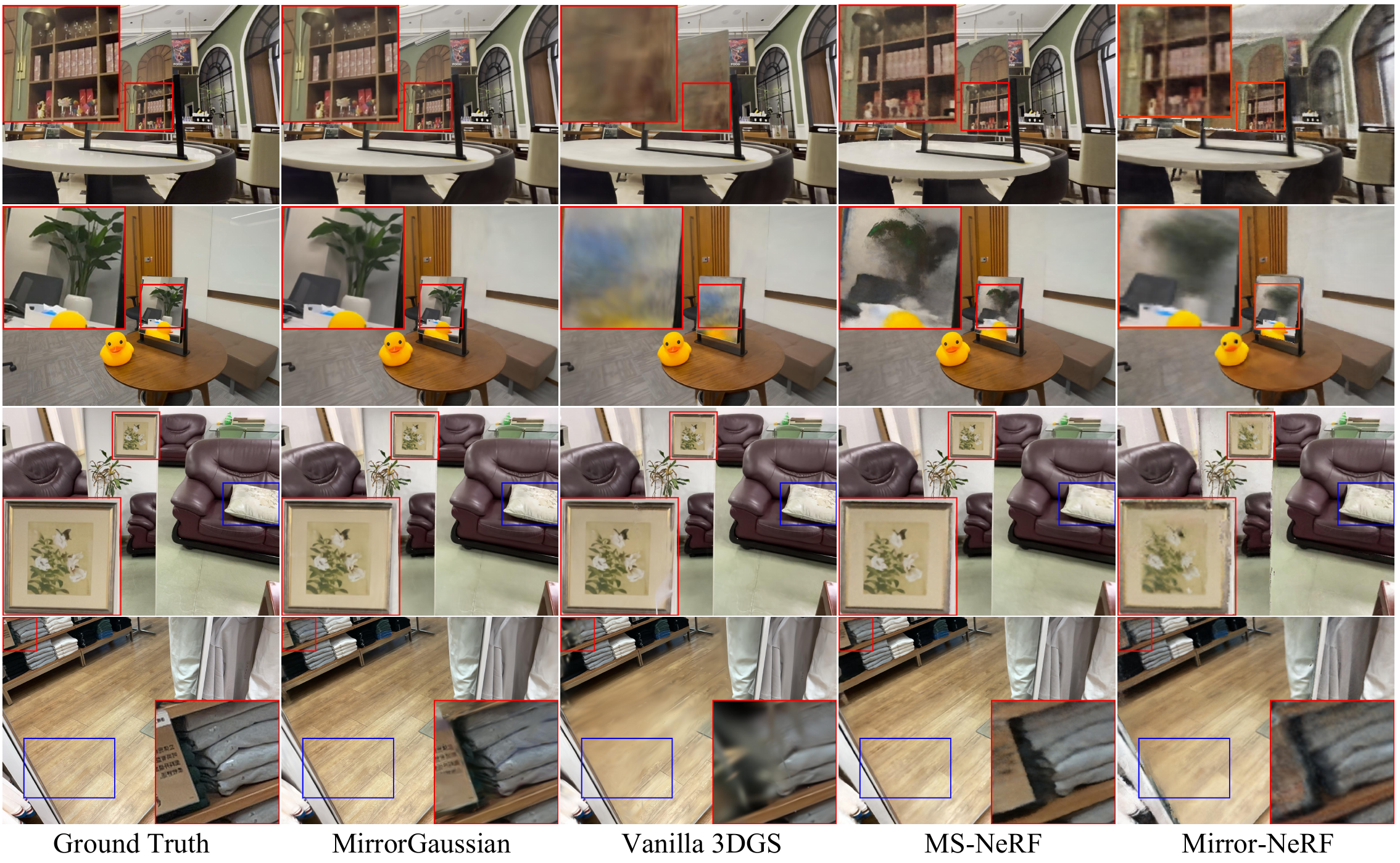}
     \caption{Reconstructed samples with different methods across various scenes. Top row: \textit{Meeting Room.} Second row: \textit{Coffee House.} Third row: \textit{Lounge.} Bottom row: \textit{Market.}}
    \label{fig: exp}
\end{figure}

\subsubsection{Compared Methods.}
We compare MirrorGaussian with the most relevant state-of-the-art methods: vanilla 3DGS\cite{kerbl20233d}, MS-NeRF\cite{yin2023multi}, Mirror-NeRF\cite{zeng2023mirror}, and Ref-NeRF\cite{verbin2022ref}. 
For vanilla 3DGS, we optimize it on four scenes under the same setting described in its original paper~\cite{kerbl20233d}.
For MS-NeRF, we choose Mip-NeRF 360~\cite{barron2022mip} based model with eight sub-spaces that shows the best quality.
For Mirror-NeRF, we use the iNGP~\cite{muller2022instant} based model for training efficiency.

\subsubsection{Datasets.}
The experiments are conducted on four real-world scenes with mirrors: \textit{Lounge} and \textit{Market} from the Mirror-NeRF dataset~\cite{zeng2023mirror}, and \textit{Meeting-Room} and \textit{Coffee-House} captured by ourselves. 
\textit{Lounge} and \textit{Market} only include forward-facing views of the mirror and omit viewpoints capturing the back. \textit{Meeting-Room} and \textit{Coffee-House} capture 360-degree views of the mirror. These two scenes are captured with an iPhone 15 Pro. We use COLMAP~\cite{schonberger2016structure} to estimate camera poses and SAM~\cite{kirillov2023segment} to segment the mirror masks.
Every 8th image is preserved for testing.
%across all scenes.

\subsubsection{Metrics.}
We perform quantitative evaluation by resizing the images to $1.6$ megapixels following~\cite{kerbl20233d} except for Mirror-NeRF.
% is performed under the resolution of $1600 \times 1200$ except for Mirror-NeRF. 
% \tx{$1600 \times 900 | 1200 $} 
We find that Mirror-NeRF is difficult to converge at high resolution, so we keep its original setting with width of $480$ pixels.
We compare all the methods using three metrics: SSIM, PSNR, and AlexNet-based LPIPS. We also report the rendering speed at the image size of $1.6$ megapixels, and the overall optimization time. %(report training time, memory)

% We extensively evaluates our method across a diverse collection of 10 scenes incorporating mirrors, including both real and synthetic scenes sourced from public and newly captured datasets. 

% In existing public datasets, real-world scenes include: \textit{Lounge}, \textit{Market} and \textit{Discussion Room} from Mirror-NeRF \cite{zeng2023mirror}, \textit{Mirror} from NeRF-ReN \cite{guo2022nerfren}, and synthetic scenes include \textit{Scene03} from MS-NeRF \cite{yin2023multi}, \textit{Livingroom}, \textit{Office} and \textit{Washroom} from Mirror-NeRF \cite{zeng2023mirror}.

% \subsection{Decomposition}
% composition visualization(including subspace)

\begin{table}[tb]
\caption{Quantitative evaluation of our method compared to previous works on four real-world scenes containing mirrors. We report SSIM $\uparrow$, PSNR $\uparrow$ and LPIPS $\downarrow$, as well as training time and FPS on test views. The \textbf{best} and \underline{second best} results are highlighted.}
  \centering
  \label{tab:table1}
\resizebox{\linewidth}{!}{
\begin{tabular}{c|ccc|ccc|ccc|ccc|cc}
\toprule
Scene                          & \multicolumn{3}{c|}{\textit{Coffee-House}} & \multicolumn{3}{c|}{\textit{Meeting-Room}} & \multicolumn{3}{c|}{\textit{Market}} & \multicolumn{3}{c|}{\textit{Lounge}}&\multicolumn{2}{c}{\textit{Average (Avg.)}} \\
\midrule
Metrics
&SSIM&PSNR&LPIPS
&SSIM&PSNR&LPIPS
&SSIM&PSNR&LPIPS
&SSIM&PSNR&LPIPS&Avg. FPS&Avg. Time \\
\midrule
MS-NeRF\cite{yin2023multi}
&0.901&\underline{26.21}&0.111
&0.910&\textbf{31.09}&0.129
&0.685&26.25&0.275
&0.921&\textbf{31.20}&0.126&0.048&14h57m \\
Mirror-NeRF\cite{zeng2023mirror}
&0.847&23.92&0.201
&0.889&27.43&0.145
&0.809&27.49&0.146
&0.917&29.38&0.184&0.468&18h23m \\
% NeRFReN\cite{guo2022nerfren}
% &-&-&-&-&-
% &-&-&-&-&-
% &0.846&26.63&0.205&- &-
% &-&-&-&-&- \\
Ref-NeRF\cite{verbin2022ref}
&0.823&23.82&0.273
&0.872&29.01&0.206
&0.560&23.75&0.522
&0.878&28.49&0.194&0.197&9h48m \\
Vanilla 3DGS\cite{kerbl20233d}
&\underline{0.921}&25.83&\underline{0.086}
&\underline{0.939}&28.82&\underline{0.073}
&\underline{0.832}&\underline{28.29}&\underline{0.127}
&\underline{0.927}&30.10&\textbf{0.107}& \textbf{271} &\textbf{47m}\\

\midrule 
\textbf{MirrorGaussian} 
&\textbf{0.927}&\textbf{26.66}&\textbf{0.068}
&\textbf{0.943}&\underline{30.63}&\textbf{0.063}
&\textbf{0.844}&\textbf{28.35}&\textbf{0.105}
&\textbf{0.928}&\underline{31.10}&\underline{0.108}&\underline{155}&\underline{1h8m}\\
\bottomrule

\end{tabular}
}
\end{table}

% \begin{table}[tb]
% \caption{Quantitative evaluation of our method compared to previous works on four real-world scenes containing mirrors. We report SSIM $\uparrow$, PSNR $\uparrow$ and LPIPS $\downarrow$, as well as training time and FPS on test views. The best and second best results are highlighted.}
%   \centering
%   \label{tab:table1}
% \resizebox{\linewidth}{!}{
% \begin{tabular}{c|ccccc|ccccc|ccccc|ccccc}
% \toprule
% Scene                          & \multicolumn{5}{c|}{\textit{Meeting-Room}} & \multicolumn{5}{c|}{\textit{Coffee-House}} & \multicolumn{5}{c|}{\textit{Lounge}} & \multicolumn{5}{c}{\textit{Market}} \\
% \midrule
% Metrics
% &SSIM&PSNR&LPIPS&FPS&Time
% &SSIM&PSNR&LPIPS&FPS&Time
% &SSIM&PSNR&LPIPS&FPS&Time 
% &SSIM&PSNR&LPIPS&FPS&Time \\
% \midrule
% MS-NeRF\cite{yin2023multi}
% &0.910&31.09&0.129&0.012&-
% &0.901&26.21&0.111&-&-
% &0.921&31.20&0.126&-&-
% &0.685&26.25&0.275&0.008&31h24m \\
% Mirror-NeRF\cite{zeng2023mirror}
% &-&-&-&0.030&-
% &-&-&-&-&-
% &0.868&27.94&0.169&0.048&-
% &-&-&-&-&- \\
% % NeRFReN\cite{guo2022nerfren}
% % &-&-&-&-&-
% % &-&-&-&-&-
% % &0.846&26.63&0.205&- &-
% % &-&-&-&-&- \\
% Ref-NeRF\cite{verbin2022ref}
% &0.872&29.01&0.206&0.280&8h51m
% &0.823&23.82&0.273&0.180&10h16m
% &0.878&28.49&0.194&0.131&10h17m
% &-&-&-&-&- \\
% Vanilla 3DGS\cite{kerbl20233d}
% &0.939&28.82&0.073&-&40m
% &0.921&25.83&0.086&-&43m
% &0.928&30.10&0.107&-&51m
% &0.832&28.29&0.127&-&1h15m\\

% \midrule
% \textbf{MirrorGaussian} 
% &0.943&30.63&0.063&-&1h3m
% &0.927&26.74&0.068&-&1h8m
% &0.929&30.80&0.108&-&1h12m
% &0.844&28.35&0.105&-&1h54m\\
% \bottomrule

% \end{tabular}
% }
% \end{table}

\subsection{Result Analysis}
\subsubsection{Rendering Quality.}
We report the SSIM, PSNR, and LPIPS metrics in each scene in \cref{tab:table1}, and also show visual comparisons in \cref{fig: exp}. The image qualities of the NeRF-based methods are relatively low, and the reflections in the parts of mirrors are even more blurry and short of details. Vanilla 3DGS produces extremely blurry mirror surfaces when the dataset contains views behind the mirror (see the first two rows in Fig.~\ref{fig: exp}). For the forward-facing scenes (the last two rows in Fig.~\ref{fig: exp}), vanilla 3DGS tends to produce point clouds behind the mirrors to model the virtual mirror space. Although the rendering from frontal perspectives is relatively good, 3DGS often gets blurry results around the mirror borders due to conflicts in different viewpoints. Note that we follow Mirror-NeRF's original setting to downscale the training images with width of $480$ pixels to avoid the extremely long training time.

\subsubsection{Training and Rendering Speeds.}
In \cref{tab:table1}, we also report the average training time and rendering speed. The training and rendering speed tests are both conducted on a single NVIDIA V100 GPU. Vanilla 3DGS and MirrorGaussian require much shorter training time, and achieves real-time rendering at high resolution. MS-NeRF builds upon Mip-NeRF 360~\cite{barron2022mip} and uses eight sub-spaces to composite the final rendering, resulting in a low rendering speed. Mirror-NeRF physically models ray reflections, and samples points on ingoing and outgoing lights recursively, thus requiring long training time. 
% MirrorGaussian extends its training from the point where Vanilla 3DGS has completed two-thirds of its training duration. It refines mirror positions and performing joint optimization up to 30,000 steps, which aligns with  the total training steps of Vanilla 3DGS, with its total training time slightly longer compared to Vanilla 3DGS. Due to our dual-rendering pipeline, the rendering speed \tx{TODO: MirrGS FPS analysis}
MirrorGaussian takes a slightly longer training time compared to vanilla 3DGS, due to the dual-rendering strategy in the final optimization stage.

% \begin{table}[tb]
% \caption{Quantitative evaluation of our method compared to previous works on four real-world scenes containing mirrors. We report training time and FPS.}
%   \centering
%   \label{tab:table2}
% \resizebox{\linewidth}{!}{
% \begin{tabular}{c|cc|cc|cc|cc}
% \toprule
% Scene                          & \multicolumn{2}{c|}{\textit{Meeting-Room}} & \multicolumn{2}{c|}{\textit{Coffee-House}} & \multicolumn{2}{c|}{\textit{Lounge}} & \multicolumn{2}{c}{\textit{Market}} \\
% \midrule
% Metrics
% &FPS&Training Time
% &FPS&Training Time
% &FPS&Training Time 
% &FPS&Training Time \\
% \midrule
% MS-NeRF\cite{yin2023multi}
% &0.012&-
% &-&-
% &-&-
% &0.008&- \\
% Mirror-NeRF\cite{zeng2023mirror}
% &0.030&-
% &-&-
% &0.048&-
% &-&- \\
% NeRFReN\cite{guo2022nerfren}
% &-&-
% &-&-
% &-&-
% &-&- \\
% Ref-NeRF\cite{verbin2022ref}
% &0.280&8h51m
% &0.180&10h16m
% &0.131&10h17m
% &-&- \\
% Vanilla 3DGS\cite{kerbl20233d}
% &-&-
% &-&43m
% &-&-
% &-&1h15m \\

% \midrule
% \textbf{MirrorGaussian} 
% &-&-
% &-&1h12m
% &-&-
% &-&1h54m \\
% \bottomrule

% \end{tabular}
% }
% \end{table}

\subsection{Ablation Studies}
\begin{figure}[tb]
    \centering
    \includegraphics[width=\textwidth]{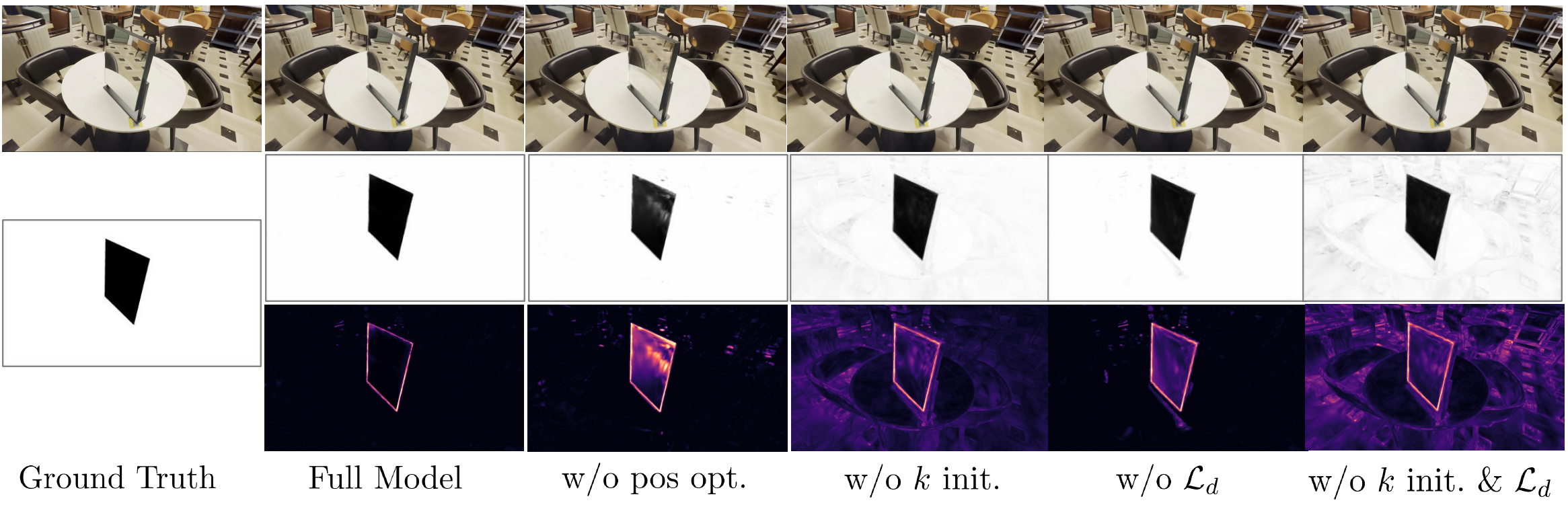}
     \caption{Ablation studies. Bottom row: FLIP error map~\cite{Andersson2020} visualization between the rendered mirror masks and the ground truth mask.}
    \label{fig: ablation}
\end{figure}

We perform ablation studies on the \textit{Coffee-House} scene to evaluate different aspects of MirrorGaussian, as illustrated in \cref{tab:ablation} and \cref{fig: ablation}.

\begin{table}[tb]
% \caption{Ablation studies on optimizing the mirror equation and strategies for mirror mask optimization.
%   }
\caption{Ablation studies on the mirror equation and mask optimization.
  }
  \centering
  \label{tab:ablation}
\resizebox{0.7\linewidth}{!}{\begin{tabular}{lccl}
\toprule
Model setting                             & SSIM$\uparrow$ & PSNR$\uparrow$ & LPIPS$\downarrow$ \\
\midrule
1) w/o mirror equation optimization          &  0.919    &  25.84    &   0.082             \\
2) w/o distance-aware initialization of \textit{k} \quad \quad &  0.926    &   26.42   &      0.071          \\
3) w/o distance loss                      &  0.926    &   26.59   &     \textbf{0.068}          \\
4) w/o both 2) and 3)                       & 0.921     & 25.95     &    0.077            \\
\midrule
\textbf{Full model}                       &  \textbf{0.927}&\textbf{26.74}&\textbf{0.068}        \\
\bottomrule
\end{tabular}}
\end{table}

\subsubsection{Mirror Equation Optimization.}
We optimize the mirror plane equation from a rough estimation in the second and final optimization stages. Without this optimization, reflecting the 3D Gaussians $P_r$ across the mirror plane $\mathcal{P}$ results in a shifted rendering in the mirror compared to the ground truth image $I_{gt}$. Consequently, the rendering results are blurry as shown in the third column of \cref{fig: ablation}.%, accompanied with some floaters as shown in the third column of \cref{fig: ablation}.

\subsubsection{Strategies for Mask Generation.}
After optimizing the mirror equation, we initialize the mirror label $k$ based on the distance of the 3D Gaussian to the mirror plane. 
% We set $k$ to $1$ when the distance exceeds a threshold, indicating that the point is not on the mirror surface. For the remaining points, $k$ is randomly assigned between $0$ and $1$. 
This initialization accelerates the 3D mirror mask optimization and thus facilitates the optimization for colors.
Besides, the distance loss $\mathcal{L}_d$ encourages mirror points to distribute over the mirror surface.
% Both strategies serve a similar purpose to facilitate mask generation in novel views. 
When the distance-aware $k$ initialization is utilized, the effects of the distance loss is not notably significant, but a much worse mask $M_r$ is produced without both. 

\subsection{Scene Editing}

\subsubsection{Adding New Objects.}
MirrorGaussian has the capability to seamlessly integrate new objects into a scene, thanks to our explict point-cloud-based representation. 
%The incorporation of any Gaussian point into the point cloud ensures instant compatibility with the pre-trained mirror within the scene. Consequently, a collection of compact 3D Gaussian points, meticulously trained to resemble an object, can effortlessly join a scene, with the added benefit of having its reflected counterpart seamlessly rendered. 
The first two figures in Fig.~\ref{fig: application} illustrates the integration of a pre-trained toy bear in front of the mirror, and the mirror shows its reflected image.
% Here, the point cloud representing the bear is trained independently and subsequently imported into the scene, precisely positioned at selected coordinates. \roy{please insert diagram here, I suggest we also put the same view without the bear side by side for comparison}

% Adjust the arrangement of objects
\subsubsection{Adding New Mirrors}
\begin{figure}[tb]
    \centering
    \includegraphics[width=\textwidth]{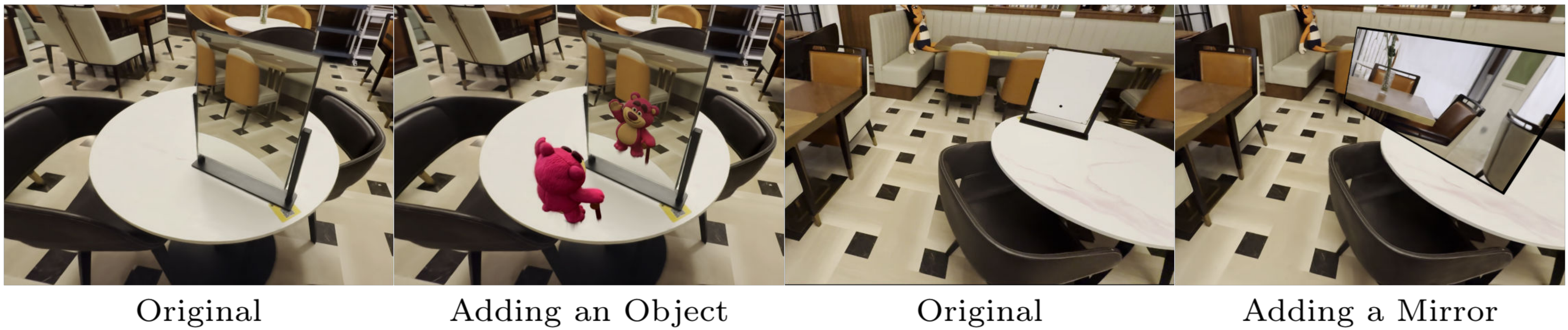}
     \caption{Illustration of scene editing. Please check our video demos in the supplementary material.}
    \label{fig: application}
\end{figure}

Likewise, we can seamlessly add a new mirror into the scene.
We generate a new mirror by distributing 3D Gaussian points evenly along the plane's surface, each sharing identical scales and rotations.
As shown in the last two figures of Fig.~\ref{fig: application}, the new mirror can correctly reflect the real-world scene.

%% file: 6.conclusion.tex
\section{Conclusion and Limitation}
% In this paper, we propose MirrorGaussian, the first method that achieves high-quality reconstruction and real-time rendering with scenes containing mirrors.
% %
% The proposed explicit point-cloud-based representation models the mirror-symmetry between the real-world space and the virtual mirror space, allowing for accurate reconstruction of reflections in mirrors.
% %
% Our rasterization-based dual-rendering strategy can further leverage it to generate high-quality images efficiently.
% %
% This translates to rapid optimization and real-time rendering, empowering various interactive applications in scene editing.

% Even though MirrorGaussian achieves promising results, there are still some limitations that remain for future work. Our method still requires preparatory mask annotations for the mirrors. Mirror segmentation models~\cite{yang2019my,tan2022mirror,huang2023symmetry} and visual-grounding-based segmentation models~\cite{ren2024grounded} can be employed to reduce workload for data preparation. Additionally, our dual-rendering strategy slightly decreases rendering speed, since we currently render and compose the real-world image and the mirror image at full resolution. This can be further optimized by focusing only on the regions containing mirrors.

In this paper, we propose MirrorGaussian, the first method that achieves high-quality reconstruction and real-time rendering for scenes containing mirrors.
The proposed explicit point-cloud-based representation utilizes the mirror symmetry between the real-world space and the virtual mirror space.
Our rasterization-based dual-rendering strategy can further leverage it to efficiently generate high-quality images in novel views, and empowers various scene editing like adding new mirrors and objects.

Even though MirrorGaussian achieves promising results in modeling mirror reflections, there are still some limitations. Our method requires mirror segmentation on input images to estimate the 3D mirror plane and mask. Additionally, our dual-rendering strategy slightly decreases the rendering speed, since we currently render and compose the real-world image and the mirror image for the whole frame. This can be further optimized by performing the dual-rendering only on the mirror region.